\documentclass{sanket}
\usepackage{cite}
\usepackage{url}
\usepackage{subcaption}
\usepackage{amsmath,amssymb}
\usepackage{float}
\usepackage{graphicx}
\usepackage{xcolor}
\usepackage{setspace}
\usepackage{booktabs}
\usepackage{upgreek}

\usepackage{array}
\newcolumntype{H}{>{\setbox0=\hbox\bgroup}c<{\egroup}@{}}

\usepackage{enumitem}
\usepackage{bm,bbm,amsthm}
\usepackage{float}
\usepackage{algorithm,algorithmic}
\usepackage{bigints}
\usepackage{hyperref}

\allowdisplaybreaks

\newcommand{\E}{\mathbb{E}}

\newcommand{\boldB}{\boldsymbol{B}}
\newcommand{\boldD}{\boldsymbol{D}}

\newcommand{\boldI}{\boldsymbol{I}}

\newcommand{\boldW}{\boldsymbol{W}}
\newcommand{\boldX}{\boldsymbol{X}}

\newcommand{\bolde}{\boldsymbol{e}}

\newcommand{\bolds}{\boldsymbol{s}}

\newcommand{\boldk}{\boldsymbol{k}}

\newcommand{\boldw}{\boldsymbol{w}}
\newcommand{\boldx}{\boldsymbol{x}}

\newcommand{\boldz}{\boldsymbol{z}}

\newcommand{\calF}{\mathcal{F}}

\newcommand{\calM}{\mathcal{M}}
\newcommand{\calN}{\mathcal{N}}

\newcommand{\indicator}{\boldsymbol{1}}

\newcommand{\bdelta}{\boldsymbol{\delta}}

\newcommand{\bsigma}{\boldsymbol{\sigma}}

\newcommand{\bTheta}{\boldsymbol{\Theta}}
\newcommand{\btheta}{\boldsymbol{\theta}}
\newcommand{\bmu}{\boldsymbol{\mu}}

\newcommand{\btau}{\boldsymbol{\tau}}

\newcommand{\prob}{\mathbb{P}}
\newcommand{\bupvarpi}{\boldsymbol{\upvarpi}}

\newcommand{\abs}[1]{\left\lvert#1\right\rvert}

\newcommand{\BlackBox}{\rule{1.5ex}{1.5ex}}  % end of proof
\ifdefined\proof
    \renewenvironment{proof}{\par\noindent{\bf Proof\ }}{\hfill\BlackBox\\[2mm]}
\else
    \newenvironment{proof}{\par\noindent{\bf Proof\ }}{\hfill\BlackBox\\[2mm]}
\fi
 
\newtheorem{theorem}{Theorem}
\newtheorem{lemma}[theorem]{Lemma} 
 
\newtheorem{remark}[theorem]{Remark}
\newtheorem{corollary}[theorem]{Corollary}
\newtheorem{definition}[theorem]{Definition}

\begin{document}
\title{Spike-and-slab shrinkage priors for structurally sparse \\Bayesian neural networks}
\author[1]{Sanket Jantre\thanks{corresponding author \\ \copyright This work has been submitted to the IEEE for possible publication.}}
\author[2]{Shrijita Bhattacharya}
\author[2]{Tapabrata Maiti}
\affil[1]{Computational Science Initiative, Brookhaven National Laboratory, Upton, NY}
\affil[2]{Department of Statistics and Probability, Michigan State University, East Lansing, MI}
\date{}

\maketitle

\begin{abstract}
\noindent%
Network complexity and computational efficiency have become increasingly significant aspects of deep learning.  Sparse deep learning addresses these challenges by recovering a sparse representation of the underlying target function by reducing heavily over-parameterized deep neural networks. Specifically, deep neural architectures compressed via structured sparsity (e.g. node sparsity) provide low latency inference, higher data throughput, and reduced energy consumption. In this paper, we explore two well-established shrinkage techniques, Lasso and Horseshoe, for model compression in Bayesian neural networks. To this end, we propose structurally sparse Bayesian neural networks which systematically prune excessive nodes with (i) Spike-and-Slab Group Lasso  (SS-GL), and (ii) Spike-and-Slab Group Horseshoe (SS-GHS) priors, and develop computationally tractable variational inference including continuous relaxation of Bernoulli variables. We establish the contraction rates of the variational posterior of our proposed models as a function of the network topology, layer-wise node cardinalities, and bounds on the network weights. We empirically demonstrate the competitive performance of our models compared to the baseline models in prediction accuracy, model compression, and inference latency.
\end{abstract}

\noindent%

\section{Introduction}
A plethora of works on sparse Bayesian neural networks (BNN) exist in the context of edge selection (e.g., \cite{Polson-Rockova-2018, Cherief-Abdellatif-2020, Bai-Guang-2020, Sun-Liang-2021, liu2019vbd}). More closely related are the works \cite{Cherief-Abdellatif-2020, Bai-Guang-2020} which use spike-and-slab prior with variational inference for a Dirac spike (allows for automated selection and offers computational gains by dropping zero weights while training \cite{Bai-Guang-2020}) and a Gaussian slab. Weight pruning although memory efficient, leads to unstructured sparsity in neural networks which insufficiently reduce the computational cost during inference at test time. This is mainly because the dimensions of the weight matrices are not reduced even if most of the weights are zero \cite{Wen-et-al-2016}. Instead, structured sparsity via node pruning - neurons in multilayer perceptrons (MLP) and channels in convolution neural networks (CNN), leads to lower computational cost at test-time due to reduced dimensions of weight matrices. Since our focus is to construct computationally compact networks, we perform node selection by use of sparsity-inducing priors.  

\begin{figure}[t!]
\centering
\begin{subfigure}[b]{0.325\textwidth}
    \centering
    \includegraphics[width=\textwidth]{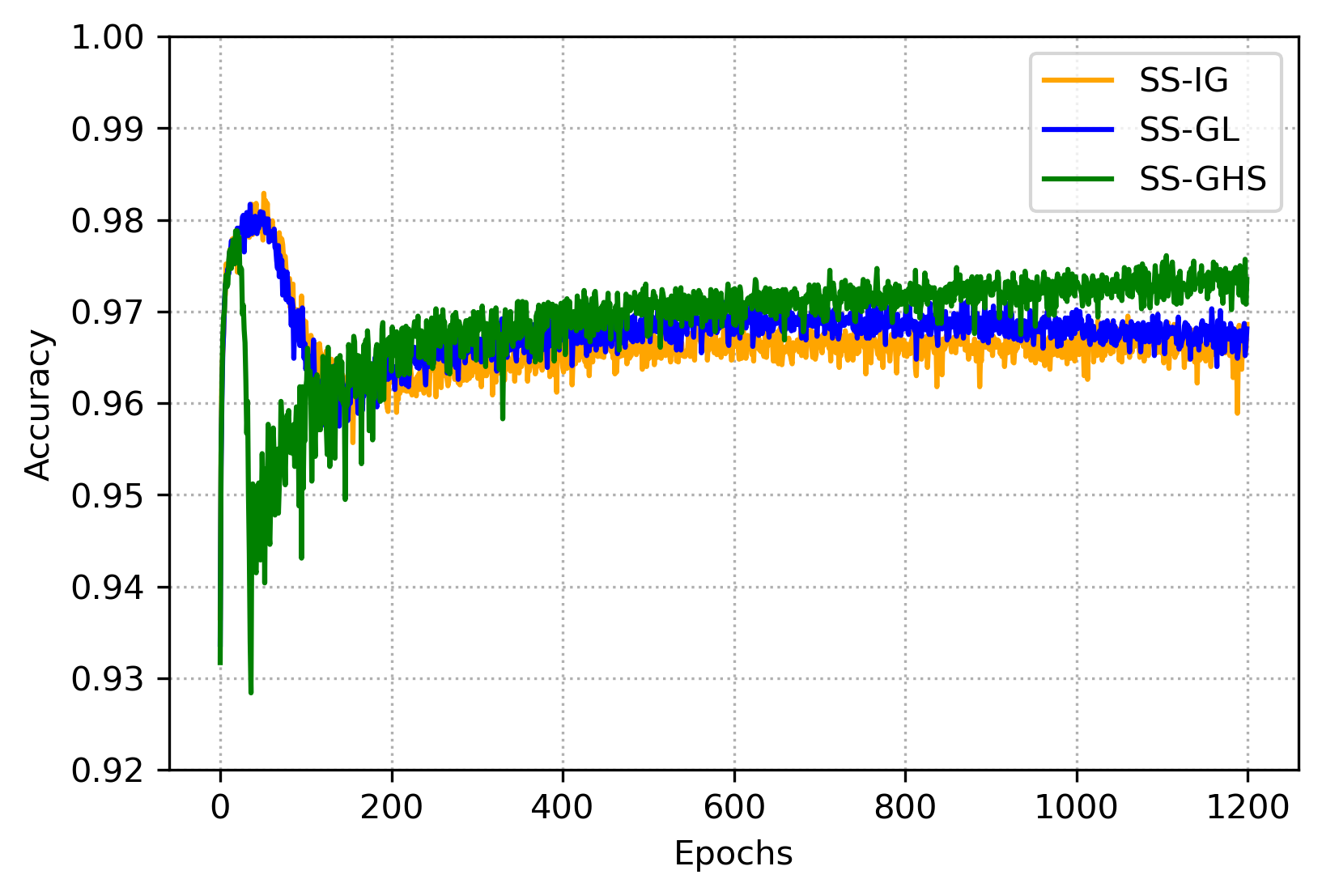} 
    \caption{{\small Prediction accuracy}}
    \label{fig:MLP-MNIST-Overall-Test-Acc-Fig1}
\end{subfigure}
\begin{subfigure}[b]{0.325\textwidth}
  \centering
  \includegraphics[width=\linewidth]{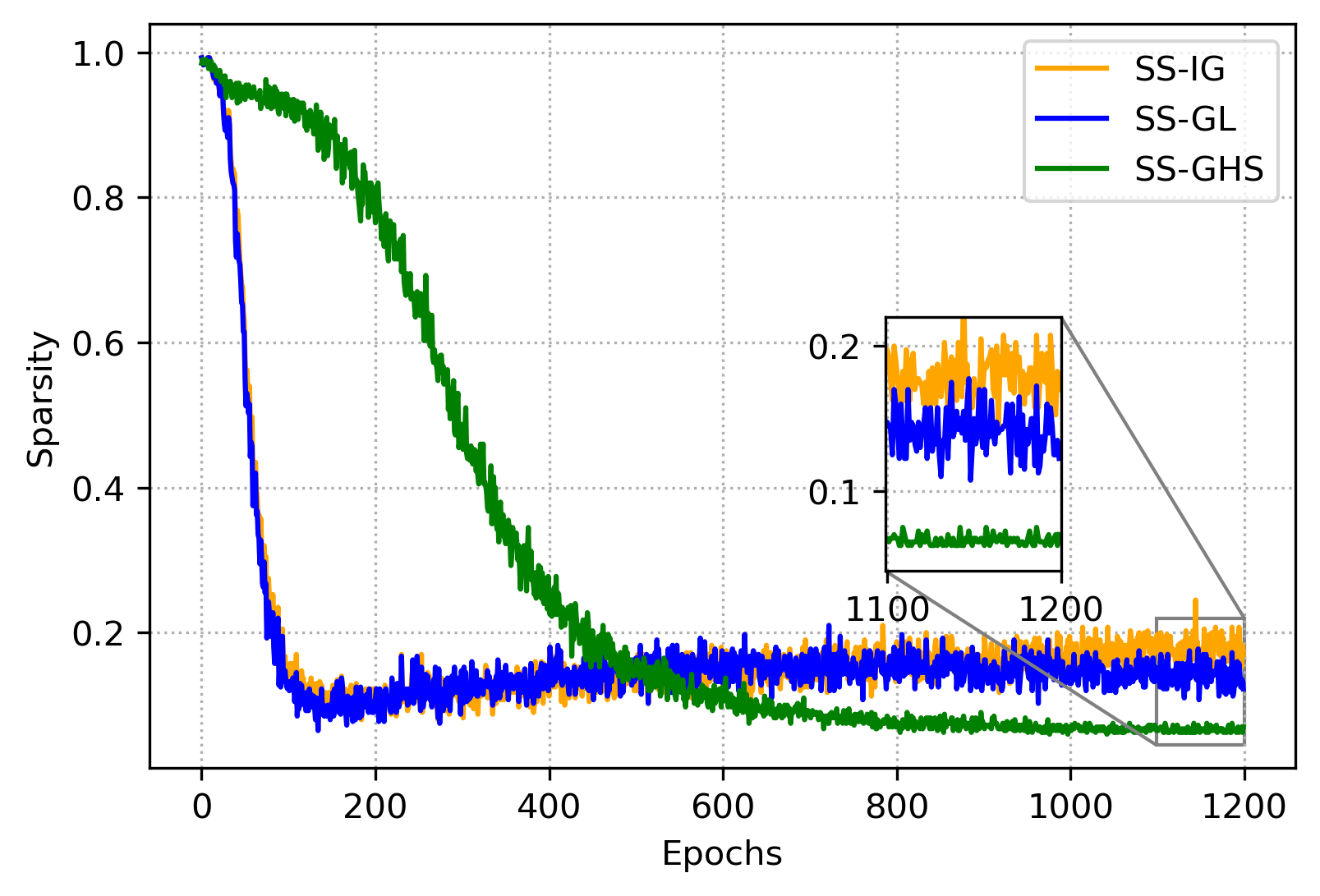}
  \caption{{\small Layer-1 node sparsity}}
  \label{fig:MLP-MNIST-layer1-node-sparsity}
\end{subfigure}
\begin{subfigure}[b]{.325\textwidth}
  \centering
  \includegraphics[width=\linewidth]{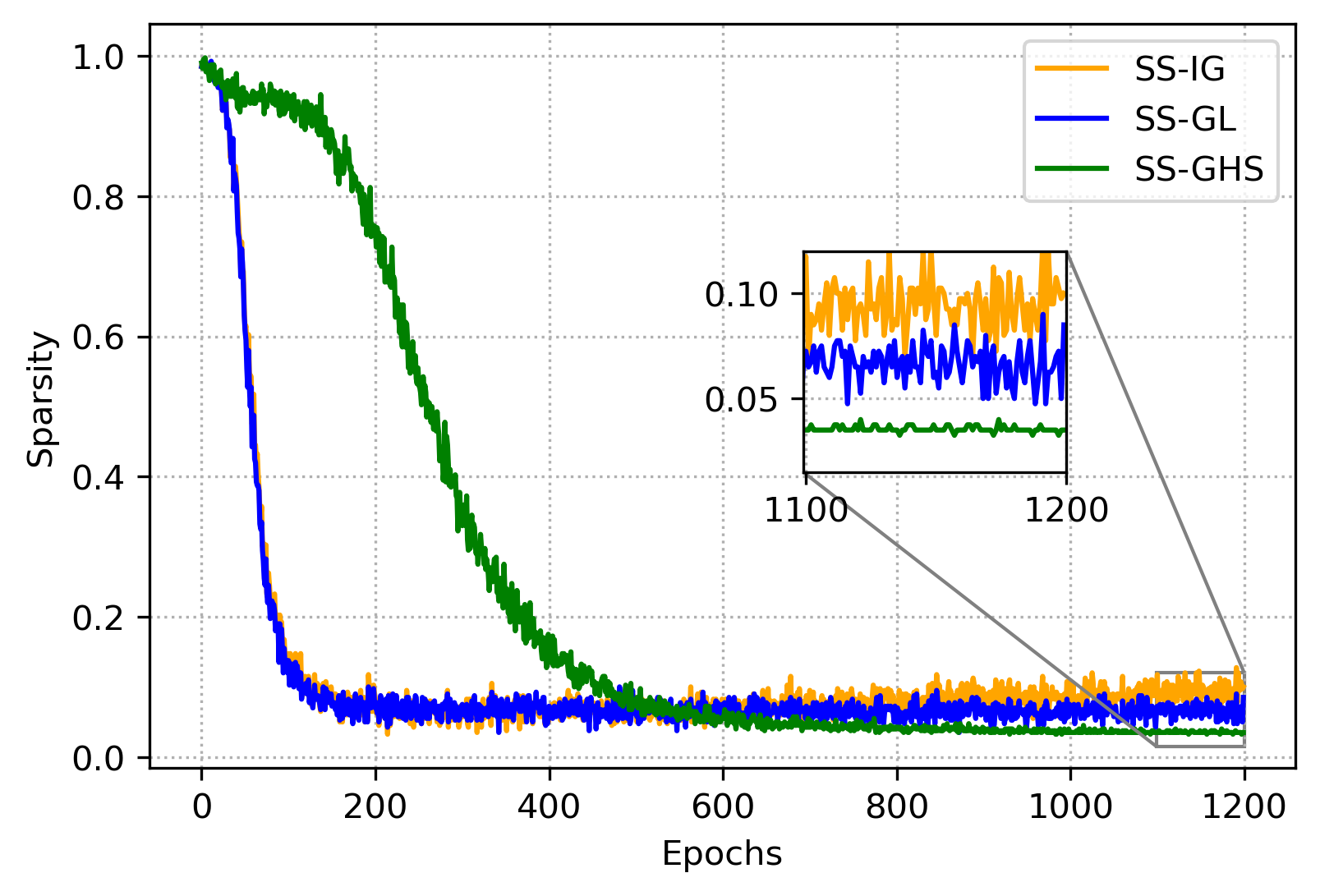}
  \caption{{\small Layer-2 node sparsity}}
  \label{fig:MLP-MNIST-layer2-node-sparsity}
\end{subfigure}
\caption{{\bf MNIST Experiment.} We demonstrate the performance of SS-GL and SS-GHS models in a 2-layer perceptron network for classifying MNIST dataset. We include closely related SS-IG model \cite{Jantre-et-al-2023} for comparison. (a) Test data prediction accuracy. (b) and (c) proportion of active nodes (node sparsity) in layer-1 and layer-2 of the network respectively. SS-GHS achieves the best predictive performance and the most compact network, as seen by the proportion of active nodes in layer-1 and layer-2.}
\label{fig:mnist-results-intro}
\end{figure}

In the context of high dimensional linear regression, two of the most popular regularization techniques are Lasso and Horseshoe \cite{Bhadra2019LassoHS}. Both Lasso and Horseshoe priors have been shown to outperform Gaussian priors in the context of Bayesian high dimensional linear regression. In this work, we employ Lasso and Horseshoe priors for neuron pruning. For an image classification task, Fig.~\ref{fig:mnist-results-intro} shows that spike-and-slab Group Lasso (SS-GL) and spike-and-slab Group Horseshoe (SS-GHS) exhibit an improvement over a simple spike-and-slab Gaussian (SS-IG) for the task of node selection.

We next discuss a few existing methods which perform the task of Bayesian node selection. \cite{Louizos-et-al-2017} proposed the use of group normal Jeffrey's prior and group Horseshoe prior for neuron pruning under variational inference (VI). \cite{Ghosh-JMLR-2018} proposed to learn structurally sparse Bayesian neural networks through node selection with Horseshoe prior via VI. These works use ad-hoc thresholding techniques for neuron pruning and do not provide any theoretical validation. The current paper resolves both issues (1) statistically principled use of spike-and-slab prior with group-shrinkage slab distributions allows automated node selection (2) derivation of the convergence rates of the variational posterior provides theoretical justification.

Currently, there exists very little work on comparative analysis of the two shrinkage priors, Lasso and Horseshoe, even in the simple scenario of high dimensional Bayesian linear regression  (see for e.g.  \cite{linear-shrinkage}).  To the best of our knowledge, there exists no work that does a thorough theoretical and numerical investigation of shrinkage priors like Lasso and Horseshoe in the context of variational BNNs.

\subsection{Related Work} 
\noindent \textbf{Sparse neural networks.} A number of approaches to neural network compression via sparsity include \cite{Cheng2018ieee, Gale2019}. Recent approaches \cite{Guo2016surgery, Molchanov-et-al-2017, Zhu2018_to_prune} in magnitude-based pruning provide high model compression rates with minimal accuracy loss. Although unstructured weight pruning  \cite{Han2015NIPS, frankle2018LOT, Bai-Guang-2020, Sun-Liang-2021} leads to a lower memory footprint of the trained model, they are difficult to map to parallel processors and can incur computational overhead during inference. Instead, a structurally compact network can be obtained using neuron pruning \cite{liu2019vbd, Alvarez2016, Ochaiai-2016, Liu_2017_ICCV, Luo_2017_ICCV, louizos2018l_0, dai2018vibnet}. However, the more closely related works include \cite{Louizos-et-al-2017} and \cite{Ghosh-JMLR-2018}. These use group shrinkage priors with variational inference to selectively shrink the collection of weights associated with a node and prune the nodes using ad-hoc thresholding criteria. 

\noindent \textbf{Approximation theory.} \cite{Schmidt-Hieber-2017} showed that estimators in nonparametric regression, sparse DNNs with ReLU activation, and wisely chosen architectures achieve the minimax estimation rates (up to log factors) under classical smoothness assumptions of the regression function. Their work forms a basis for the recent theoretical works on sparse neural networks involving weight pruning approaches under the Bayesian framework such as \cite{Polson-Rockova-2018} and \cite{Sun-Liang-2021}. \cite{Cherief-Abdellatif-2020} and \cite{Bai-Guang-2020}  develop variational posterior consistency and its contraction rates for their spike-and-slab Gaussian priors in the context of edge selection. 

\subsection{Our Contributions}
\begin{itemize}
    \item We investigate structurally sparse Bayesian neural networks with the use of two distinct spike-and-slab priors, where the slab uses shrinkage priors on the group of incoming weights (including the bias term) on the neurons: (i) {\bf S}pike-and-{\bf S}lab {\bf G}roup {\bf L}asso ({\bf SS-GL}) (ii) {\bf S}pike-and-{\bf S}lab {\bf G}roup {\bf H}orse{\bf S}hoe ({\bf SS-GHS}).
    \item We provide variational posterior consistency and optimal posterior contraction rates for SS-GL and SS-GHS under regularity assumptions of network topology along with characterization of optimal hyperparameter choice.
    \item We enunciate the numerical challenges of implementing the SS-GL and SS-GHS for varying neural network architectures on a wide class of image classification datasets. 
\end{itemize}

\section{Preliminaries}
\subsection{Bayesian Neural Networks}
Let $Y \in \mathbb{R}$, $X \in \mathcal{X}$ be two random variables with the following conditional distribution
\begin{equation}
\label{e:class}
f_0(y|\boldx)=\exp\left[h_1(\eta_0(\boldx))y+h_2(\eta_0(\boldx))+h_3(y)\right]
\end{equation}
where $\eta_0(\cdot):\mathcal{X}\rightarrow \mathbb{R}$ is a continuous function satisfying certain regularity assumptions,  $\mathcal{X}$ is usually a compact subspace of $\mathbb{R}^p$ and the functions $h_1, h_2, h_3$ are pre-determined. For $h_1(u)=u$, $h_2(u)=-u^2$, $h_3(y)=-y^2/2-\log (2\pi)/2$,  we get the  regression model with  $\sigma^2=1$.   Note, $\boldx$ is a feature vector from the marginal distribution $P_{X}$, and $y$ is the corresponding output from $Y|X=\boldx$ in \eqref{e:class}. Let $P_{X,Y}$ denote the joint distribution of $(X,Y)$. 

Let $g:\mathcal{X} \to \mathbb{R}$ be a measurable function, the  risk of  $g$ is $R(g)=\int_{\mathcal{Y}\times \mathcal{X}}L(Y,g(X))dP_{X,Y}$ for some loss function $L$. The Bayes estimator minimizes this risk \cite{friedman2001elements}.  For regression with squared error loss,
%and classification with 0-1 loss,
the optimal Bayes estimator is  $ g^*(\boldx)=\eta_0(\boldx)$
%and $g^*(\boldx)=1\{\eta_0(\boldx)\geq 0\}$ respectively. 
Practically, the Bayes estimator is not useful since the function $\eta_0(\boldx)$ is unknown. Thus, an estimator is obtained based on the training observations, $\mathcal{D}=\{(\boldx_1,y_1),...,(\boldx_n,y_n)\}$. For a good estimator, the risk approaches Bayes risk as $n \to \infty$ irrespective of  $P_X$. To this end, we use Bayesian neural networks, $\eta_{\btheta}(\boldx)$ with $\btheta$ denoting the network weights, as an approximation to $\eta_0(\boldx)$.

For $\boldx \in \mathbb{R}^p$, consider a BNN with $L$ hidden layers and $k_{1}, \cdots, k_{L}$  number of nodes in the hidden layers. Set $k_{0}=p$ and $k_{L+1}=1$ ({\it $k_{L+1}>1$ allows the generalization to $Y \in \mathbb{R}^{d}, d>1$, hence providing a  handle on multi-class classification problems.}) The total number of parameters is  $K=\prod_{l=0}^L k_{l+1} (k_l+1)$. With $\boldW_l=[\boldw^{(1)}_{l}, \boldW^{(-1)}_{l}]$ ($\boldw^{(1)}_l$ is the first column and $\boldW^{(-1)}_{l}$ is all but the first column of $\boldW_{l}$), 
\begin{equation}
\label{e:eta-x}
\eta_{\btheta}(\boldx)=\boldw^{(1)}_{L}+\boldW^{(-1)}_{L}\psi(\cdots \psi(\boldw^{(1)}_0+\boldW^{(-1)}_0 \boldx))),
\end{equation}
where $\psi$ is a nonlinear activation function, $\boldw^{(1)}_{l}$ are $k_{l+1} \times 1$ vectors and $\boldW^{(-1)}_{l}$ are  $k_{l+1} \times k_{l}$ matrices.
Approximating $\eta_0(\boldx)$ by $\eta_{\btheta}(\boldx)$, conditional probabilities of $Y|X=\boldx$ are 
\begin{equation}
\label{e:sig-x}
f_{\btheta}(y|\boldx)=\exp\left[h_1(\eta_{\btheta}(\boldx))y+h_2(\eta_{\btheta}(\boldx))+h_3(y)\right].
\end{equation}
For the data $\mathcal{D}$, we denote the likelihood function as
\begin{equation}
\label{e:lik}
P_{\btheta}^n=\prod_{i=1}^n f_{\btheta}(y_i|\boldx_i), \hspace{5mm}
P_0^n=\prod_{i=1}^n f_{0}(y_i|\boldx_i).
\end{equation} under the model and the truth respectively.

\subsection{Variational Inference}
We denote all the parameters of BNN with $(\btheta,\bupvarpi)$, where $\bupvarpi$ are the extra parameters beyond the network weights $\btheta$. The posterior density of  $(\btheta,\bupvarpi)$ given the data $\mathcal{D}$ is
\begin{equation}
\label{e:posterior}
\pi(\btheta,\bupvarpi|\mathcal{D}) = \frac{\ P_{\btheta}^n \pi(\btheta,\bupvarpi)}{\int P_{\btheta}^n d\pi(\btheta,\bupvarpi)d\btheta d\bupvarpi} = \frac{ P_{\btheta}^n \pi(\btheta,\bupvarpi)d\btheta,\bupvarpi}{m(\mathcal{D})}
\end{equation}
where $P_{\btheta}^n$ is defined in \eqref{e:lik} and  $m(\mathcal{D})$ is the marginal density of the data  free of $(\btheta,\bupvarpi)$. Let $\Pi(.|\mathcal{D})$ be the cumulative  distribution function of the probability density function in \eqref{e:posterior}.

Let $\widetilde{\pi}(\btheta)=\int \pi(\btheta,\bupvarpi) d\bupvarpi$ be the marginal density of $\btheta$ under the prior and $\widetilde{\Pi}$ be the corresponding distribution function. The marginal posterior density of $\btheta$ can be rewritten as 
\begin{equation}
\label{e:marginal-posterior}
    \widetilde{\pi}(\btheta|\mathcal{D})= \int \pi(\btheta,\bupvarpi|\mathcal{D}) d\bupvarpi=\frac{P_{\btheta}^n \widetilde{\pi}(\btheta)}{\int P_{\btheta}^n \widetilde{\pi}(\btheta)d\btheta}=\frac{P_{\btheta}^n \widetilde{\pi}(\btheta)}{m(\mathcal{D})}
\end{equation}
Let $\widetilde{\Pi}(.|\mathcal{D})$ be the distribution function for the density  in \eqref{e:marginal-posterior}.

Since the posterior density in \eqref{e:posterior} is intractable,  we employ VI to approximate it. Variational learning proceeds by inferring parameters of a distribution on the model parameters, $q$, by minimizing the Kullback-Leibler (KL) distance from the true Bayesian posterior, $\pi(.|\mathcal{D})$ (\cite{Blei_2007}, \cite{Hinton93} ) as follows
\begin{equation}
\label{e:var-posterior}
    \pi^*=\underset{q \in \mathcal{Q}}{\text{argmin}}\:\: d_{\rm KL}(q,\pi(.|\mathcal{D}))=\underset{q \in \mathcal{Q}}{\text{argmax}}\:\: \text{ELBO}(q,\pi(.|\mathcal{D}))
\end{equation}
where $d_{\rm KL}(.,.)$ denotes the KL-distance, $\mathcal{Q}$ denotes the variational family and ELBO denotes evidence lower bound. We shall use the notation $\Pi^*$ to denote the distribution function corresponding to the  optimal density $\pi*$. For a density $q(\btheta,\bupvarpi)$, the negative ELBO $(\mathcal{L})$ is given by
\begin{equation}
\label{e:neg-elbo} 
    \mathcal{L}= -\mathbb{E}_{q(\btheta,\bupvarpi)} [\log L(\btheta)] + d_{\rm KL}(q(\btheta,\bupvarpi),\pi(\btheta,\bupvarpi)),
\end{equation}
where $L(\btheta)$ is the likelihood function in \eqref{e:lik}. Finally,  let  $\widetilde{\pi}^*(\btheta)=\int \pi^*(\btheta|\bupvarpi)\pi^*(\bupvarpi) d\bupvarpi$ be the marginal variational posterior of $\btheta$ and $\widetilde{\Pi}^*$ be its corresponding probability distribution.

\section{ Structured Sparsity: Spike-and-Slab Hierarchical Priors}
For automatic node selection, we consider spike-and-slab priors with  a Dirac spike ($\delta_0$) at 0. The spike part, denoted by an indicator variable, is 0 if a node is not present in the network. Although zero-mean Gaussian are commonly used as slab distribution for spike-and-slab priors \cite{Jantre-et-al-2023}, they produce inflated predictive uncertainties, especially if used in conjunction with fully factorized variational families \cite{Ghosh-JMLR-2018}. Instead, if we consider a zero-mean Gaussian slab distribution with its scale being a random variable then the slab part of the marginal prior distribution will have heavier tails and higher mass at zero. Such hierarchical distributions in slab improve the sparsity and circumvent the inflated predictive uncertainties. For optimal layer-wise node selection, we let the prior inclusion probabilities $\lambda_l$ to depend on the layer index $l$. We assume a spike-and-slab prior and a corresponding spike-and-slab variational family. We use $N(.,.)$, $\text{Ber}(.)$, $G(.,.)$, $LN(.,.)$, $C^+(0,1)$,  $IG(.,.)$ to denote Gaussian, Bernoulli, Gamma, Log-Normal, Half Cauchy and Inverse Gamma distributions.

\subsection{ Spike-and-slab group lasso (SS-GL):}
\label{sec:ssgl-prior}
\noindent{\bfseries Prior:} Let $z_{lj}$ be the indicator for the presence of $j^{\rm th}$ node in the $l^{\rm th}$ layer. Assume the following prior
\begin{equation}
\left.
\begin{aligned}
\label{e:group-lasso-prior}
\pi(\boldw_{lj}|z_{lj}) &= (1-z_{lj})\bdelta_0 + z_{lj} N(0,\sigma_0^2 \tau_{lj}^{2} \boldI)\\
\pi(z_{lj}) &= \text{Ber}(\lambda_l)\\
\pi(\tau_{lj}^{2}|\varsigma^2) &= G(k_l/2+1, \varsigma^2/2),\quad \pi(\varsigma^2) = G(a_0,b_0)
\end{aligned}
\right\}
\end{equation}
$l=0,\cdots,L$, $j=1, \cdots, k_{l+1}$ and  $\boldw_{lj} = (w_{lj1}, \cdots, w_{lj{k_l+1}})$ is the $j^{\rm th}$ row of $W_l$ (vector of edges incident onto the $j^{\rm th}$ node in the $l^{\rm th}$ layer).  $\bdelta_0$ is a Dirac vector of dimension $k_l+1$ with all zeros and $\boldI$ is  identity matrix of dimension $(k_l+1) \times (k_l+1)$. We assume for all $j$, $z_{lj}$  follow $\text{Ber}(\lambda_l)$ to allow for a common prior on the inclusion probability of all nodes in layer $l$. Let $\lambda_L=1$ to ensure no node selection occurs in the output layer.

Finally, $\sigma_0$ is a known global constant and $\tau_{lj}$ is Gamma distributed local variable (per node) with hyperparameter $\varsigma$. \cite{SS-GL-Ghosh-2015} demonstrates why the scale mixture representation in \eqref{e:group-lasso-prior} indeed corresponds to Group Lasso. To reduce sensitivity to the hyperparameter $\varsigma^2$, the Gamma prior as in \eqref{e:group-lasso-prior} is used. \vspace{0.1in}

\noindent {\bfseries Variational family:} Consider the following variational family
\begin{equation}
\left.
\begin{aligned}
\label{e:group-lasso-var}
q(\boldw_{lj}|&z_{lj}) =
(1-z_{lj})\bdelta_0+ z_{lj} N(\bmu_{lj},\text{diag}(\bsigma^2_{lj}))\\
q(&z_{lj}) =\text{Ber}(\gamma_{lj})\\
\hspace{-2mm} q(\tau_{lj}^{2}) = \:
&LN(\mu^{\{\tau\}}_{lj},{\sigma^{\{\tau\}2}_{lj}}),\: q(\varsigma^2) = LN(\mu^{\{\varsigma\}},{\sigma^{\{\varsigma\}2}})
\end{aligned}
\right\}
\end{equation}
where $\bmu_{lj}=(\mu_{lj1},\dots,\mu_{lj{k_l+1}})$ and $\bsigma^2_{lj}=(\sigma^2_{lj1},\dots,\sigma^2_{lj{k_l+1}})$ are the variational mean and standard deviation for slab $q(\boldw_{lj}|z_{lj})$ and $\text{diag}(\bsigma^2_{lj})$ is diagonal matrix with entries  $\bsigma^2_{lj}$. The choice of a fully factorized Gaussian family is motivated from \cite{Blei2017, blundell2015weight}. Variational inclusion probability under $q(z_{lj})$ is $\gamma_{lj}$ ($\gamma_{Lj}=1$ to avoid node selection in the output layer). Finally, $\mu^{\{\tau\}}_{lj}$ and $\sigma^{\{\tau\}2}_{lj}$ are variational mean and standard deviation of $q(\log \tau_{lj}^{2})$ and $\mu^{\{\varsigma\}}$ and $\sigma^{\{\varsigma\}2}$ are variational mean and standard deviation of $q(\log \varsigma^{2})$.

 The spike-and-slab distribution in the variational family ensures that the weights follow spike-and-slab structure allowing for exact node sparsity. The weight distributions conditioned on the node indicators are all independent (see \cite{titsias2011spike, liu2019NMF, Ray-Szabo-2021,  Jantre-et-al-2023} which use similar variational families). The use of Log-Normal family to approximate the Gamma distributed $\tau_{lj}^{2}$ and $\varsigma^2$ is to facilitate numerical optimization (the statistical validity is provided in Section \ref{sec:theory}). Note  the independence of $\boldw_{lj}$ (global variables) and the parameters $\tau_{lj}^2$  and $\varsigma^2$ in the variational family. Treating $\tau_{lj}^2$ and $\varsigma^2$ as hidden latent variables, we refer to \cite{Blei2017}, \cite{YD2019}, \cite{liu2019vbd} which suggest the independence of global and latent variables in the variational family. Note, one could have used a variational family mimicking the conditional dependence structure of $\boldw_{lj}$ and $\tau_{lj}$ in the prior \eqref{e:group-lasso-prior}. Since our preliminary experiments revealed the suboptimal performance of such a variational family, we stuck to the choice in \eqref{e:group-lasso-var}.

\vspace{0.1in}
\noindent {\bfseries ELBO:} For SS-GL, $\btheta$ includes all weight parameters $\boldw_{lj}$ and $\bupvarpi$ includes all latent variables $z_{lj}$, $\tau_{lj}$ and $\varsigma^2$ (see Appendix A in Supplement for derivation of $\mathcal{L}$ in \eqref{e:neg-elbo} for SS-GL).

\subsection{ Spike-and-slab group horseshoe (SS-GHS):}
\label{sec:ssghs-prior}
\noindent {\bfseries Prior:}  We consider the regularized version of group horseshoe (GHS) \cite{Piironen-Vehtari-2017} for the slab to circumvent the numerical stability issues of unregularized group horseshoe. Assume 
\begin{equation}
\left.
\begin{aligned}
\label{e:group-horse-shoe-prior}
\pi(\boldw_{lj}|z_{lj})&=  
(1-z_{lj})\bdelta_0 + z_{lj} N(0,\sigma_0^2 \widetilde{\tau}_{lj}^{2} \zeta^2 \boldI) \\
\pi(z_{lj}) &= \text{Ber}(\lambda_l)\\
\pi(\tau_{lj}) &=
 C^+(0,1),\:\: \pi(\zeta) = C^+(0,d_0)
\end{aligned}
\right\}
\end{equation}
where $\widetilde{\tau}_{lj}^{2} = (c^2_{\rm reg}\tau_{lj}^{2})/(c^2_{\rm reg}+\tau_{lj}^{2}\zeta^2)$ is a node wise varying local scale parameter,  $\zeta^2$ is a varying global scale parameter with hyperparameter  $d_0$ and $\lambda_l$, $\sigma_0^2$ are as defined in SS-GL. 

When weights strongly shrink towards 0, then $\tau_{lj}^{2}\zeta^2 \ll c^2_{\rm reg}$ and $\widetilde{\tau}_{lj}^{2} \to \tau_{lj}^{2}\zeta^2$ which leads to unregularized version of GHS. When weights are away from 0, $\tau_{lj}^{2}\zeta^2$ is large, then  $\tau_{lj}^{2}\zeta^2 \gg c^2_{\rm reg}$ and $\widetilde{\tau}_{lj}^{2} \to c^2_{\rm reg}$ for a constant $c^2_{\rm reg}$.  For these weights, the regularized GHS in the slab follows $N(0,\sigma_0^2 c^2_{\rm reg}\boldI)$, thereby thinning out the heavy tails of horseshoe.  

Instead of dealing with half-Cauchy distribution, we reparameterize half-Cauchy using Gamma and Inverse Gamma \cite{Louizos-et-al-2017}. This allows us to compute the KL-divergence between prior $\pi(\tau)$ and log-normal variational posterior $q(\tau)$ in closed form. Let $\tau^2=\tilde{\beta}\tilde{\alpha}$ where  $\tilde{\beta} \sim IG(1/2,1), \tilde{\alpha} \sim G(1/2,k^2)$, then $\tau \sim C^+(0,k)$. Using this result, we rewrite the prior in \eqref{e:group-horse-shoe-prior} as
\begin{equation}
\left.
\begin{aligned}
\label{e:group-horse-shoe-prior-split}
\pi(\boldw_{lj}|z_{lj}) &=(1-z_{lj})\bdelta_0 + z_{lj} N(0,\sigma_0^2 \widetilde{\tau}_{lj}^{2} s^2 \boldI)\\
\pi(z_{lj}) &= \text{Ber}(\lambda_l)\\
\pi(\beta_{lj}) &= IG(1/2,1), \quad \pi(\alpha_{lj}) = G(1/2,1)\\
\pi(\zeta_b) &= IG(1/2,1), \quad \pi(\zeta_a) = G(1/2,d_0^2)
\end{aligned}
\right\}
\end{equation}
where $\widetilde{\tau}_{lj}^{2} = (c^2_{\rm reg}\beta_{lj}\alpha_{lj})(c^2_{\rm reg}+\beta_{lj}\alpha_{lj}\zeta_a\zeta_b)^{-1}$ and $\zeta^2=\zeta_a\zeta_b$.

\begin{algorithm}[b!]
\caption{SS-GL and SS-GHS Bayesian neural networks}
\label{alg-SS-GL-GHS}
\begin{algorithmic}
\setlength{\leftskip}{-1em}
\setlength{\labelsep}{0.5em}
\STATE {\bfseries Inputs:} training dataset, network architecture, and optimizer tuning parameters.
\STATE \textit{Model inputs:} prior parameters: $(\btheta, \bupvarpi)$.
\STATE \textit{Variational inputs:} number of Monte Carlo samples $S$.
\STATE {\bfseries Output:} Variational parameter estimates of network weights, scales, and sparsity.
\STATE {\bfseries Method:} Set initial values of variational parameters.
\REPEAT
\setlength{\leftskip}{-1em}
\setlength{\labelsep}{0.5em}
    \STATE Generate $S$ samples of $ \boldw_{lj}, z_{lj}$ and $\tau_{lj}^2, \varsigma^2$ in SS-GL and $\beta_{lj}, \alpha_{lj}, \zeta_a, \zeta_b$ in SS-GHS.
    \STATE Use $\boldw_{lj}, z_{lj}$ and $\tau_{lj}^2, \varsigma^2$ in SS-GL and $\beta_{lj}, \alpha_{lj}, \zeta_a, \zeta_b$ in SS-GHS to compute $\mathcal{L}$ in forward pass
    \STATE Use $\boldw_{lj},\tilde{z}_{lj}$  and $\tau_{lj}^2, \varsigma^2$ in SS-GL and $\beta_{lj}, \alpha_{lj}, \zeta_a, \zeta_b$ in SS-GHS to compute $\nabla\mathcal{L}$ in backward pass
    \STATE Update the variational parameters with gradient of loss using stochastic gradient descent algorithm (e.g. \cite{sutskever-SGD-2013})
\UNTIL{change in ELBO $> \epsilon$}
\end{algorithmic}
\end{algorithm}

\vspace{0.1in}
\noindent {\bfseries Variational family:} Consider the following variational family
\begin{equation}
\left.
\begin{aligned}
\label{e:group-horse-shoe-var}
(\boldw_{lj}|&z_{lj}) = (1-z_{lj})\bdelta_0+ z_{lj} N(\bmu_{lj},\text{diag}(\bsigma^2_{lj}))\\ 
q(&z_{lj}) = \text{Ber}(\gamma_{lj})\\
\hspace{-3.9mm} q(\beta_{lj}) = \: &LN(\mu^{\{\beta\}}_{lj}, {\sigma^{\{\beta\}2}_{lj}}), q(\alpha_{lj}) = LN(\mu^{\{\alpha\}}_{lj}, {\sigma^{\{\alpha\}2}_{lj}}) \\
\hspace{-3.9mm} q(\zeta_b) \!=\! L&N(\mu^{\{\zeta_b\}}, {\sigma^{\{\zeta_b\}2}}), q(\zeta_a) \!=\! LN(\mu^{\{\zeta_a\}}, {\sigma^{\{\zeta_a\}2}})  
\end{aligned}
\right\}
\end{equation}
Note, $\bmu_{lj}$, $\bsigma_{lj}^2$, and $\gamma_{lj}$ follow the same definition as in SS-GL. However, $(\mu^{\{\beta\}}_{lj}, {\sigma^{\{\beta\}2}_{lj}})$, $(\mu^{\{\alpha\}}_{lj}, {\sigma^{\{\alpha\}2}_{lj}})$, $(\mu^{\{\zeta_b\}}, {\sigma^{\{\zeta_b\}2}})$, and  $(\mu^{\{\zeta_a\}}, {\sigma^{\{\zeta_a\}2}})$ represent variational mean and standard deviation parameters of the Gaussian distributions  -- $q(\log \beta_{lj}), q(\log \alpha_{lj}), q(\log \zeta_b)$ and $q(\log \zeta_a)$.

Similar to SS-GL, we assume the weight distributions conditioned on the node indicators are independent. We also assume the independence of $\boldw_{lj}$ and the scale indexing parameters $\beta_{lj}$, $\alpha_{lj}$, $\zeta_a$ and $\zeta_b$. All scale indexing parameters are treated as hidden latent variables and the weights $\boldw_{lj}$ are treated as global variables. The independence of the latent and global parameters in the variational family is motivated by works of \cite{Blei2017, YD2019, liu2019vbd}. Lastly, the use Log-Normal family to approximate Gamma and Inverse-Gamma variables facilitates numerical optimization (statistical validity is provided in Section \ref{sec:theory}).

\vspace{0.1in}
\noindent{\bfseries ELBO:} For SS-GHS, $\btheta$ is same as SS-GL and $\bupvarpi$ includes the latent variables $z_{lj}$, $\beta_{lj}$, $\alpha_{lj}$, $\zeta_a$ and $\zeta_b$ (see Appendix B in Supplement for derivation of $\mathcal{L}$ in \eqref{e:neg-elbo} for SS-GHS).

\subsection{ Algorithm and Computational Details}
We minimize the loss $\mathcal{L}$ for SS-GL and SS-GHS models by recursively sampling from their corresponding variational posterior. The Gaussian variational approximations, $N(\bmu_{lj},\text{diag}(\bsigma^2_{lj}))$, are reparameterized as $\bmu_{lj}+\bsigma_{lj}\odot \bolde_{lj}$ for $\bolde_{lj}\sim N(0,\boldI)$ ($\odot$ denotes entry-wise (Hadamard) product).

\vspace{0.1in}
\noindent {\bf Continuous Relaxation.} The discrete variables $\boldz$ are replaced by their continuous relaxation to circumvent the non differentiablility in $\mathcal{L}$ \cite{Jang2017Gumbel,Maddison2017Contrelax}. The Gumbel-softmax (GS) distribution is used for continuous relaxation, where $q(z_{lj}) \sim \text{Ber}(\gamma_{lj})$ is approximated by $q(\tilde{z}_{lj}) \sim {\rm GS}(\gamma_{lj},\text{temp})$, 
$\tilde{z}_{lj} = 1/(1+\exp(-\eta_{lj}/\text{temp}))$,  $\eta_{lj} = \log(\gamma_{lj}/(1-\gamma_{lj})) + \log (u_{lj}/(1-u_{lj}))$, $ u_{lj} \sim U(0,1)$ and the temperature parameter, temp = 0.5 \cite{Jantre-et-al-2023}. Using $\tilde{z}_{lj}$ in backward pass eases gradient calculation. Using $z_{lj}$ in forward pass gives exact node sparsity.

\section{Structurally Sparse BNNs: Theory}
\label{sec:theory}
Let $\widetilde{\boldw}_l=(||\boldw_{l1}||_2, \cdots, ||\boldw_{lk_{l+1}}||_2)$ be the vector of $L_2$ norms of rows of $W_l$. We rely on the following assumption on the NN architecture: (1) the number of nodes can vary with the layer index (2) the best sparse neural network approximation to the function $\eta_0$ (i) can have varying number of nodes in each layer, that is layer-wise node sparsity $\bolds=(s_1, \cdots, s_L)$ (ii) can have varying bounds per layer, $\boldB=(B_1, \cdots, B_L)$, on the $L_2$ norm of the incoming weights onto each node. 

Let $f_0$ and $f_{\btheta}$ be the joint density of  $(y_i, \boldx_i)_{i=1}^n$ under the truth and the model respectively.  For this section, we assume $\boldX \sim U[0,1]^p$ (see also  \cite{Polson-Rockova-2018, Sun-Liang-2021}), implying that $f_0(\boldx)=f_{\btheta}(\boldx)=1$ and the joint distribution of $(y_i,\boldx_i)_{i=1}^n$ depends only on conditional distribution of $Y|X=\boldx$, i.e, $P_0=f_0(y|\boldx)f_0(\boldx)=f_0(y|\boldx)$ and similarly $P_{\btheta}=f_{\btheta}(y|\boldx)$.

Consider a sieve of neural networks (check definition A.1 in  \cite{Jantre-et-al-2023})
$$\mathcal{F}(L,\boldk,\bolds,\boldB)=\left\{\eta_{\btheta} \in \eqref{e:eta-x}: ||\widetilde{\boldw}_l||_0 \leq s_l, ||\widetilde{\boldw}_l||_\infty \leq B_l  \right\}$$
Define the distance between  true function $\eta_0$ and the sieve as
 \begin{equation}
 \label{e:xi-def}
     \xi = \text{min}_{\eta_{\btheta} \in \mathcal{F}(L,\boldk,\bolds, \boldB)}||\eta_{\btheta}-\eta_0||^2_\infty.
 \end{equation}
Next, consider any sequence $\epsilon_n \to 0$. In the following two theorems, we establish the variational posterior consistency of $\widetilde{\Pi}^*$, i.e the variational posterior concentrates in shrinking $\epsilon_n$-Hellinger neighborhoods (Appendix~\ref{AppendixA} Definition 1) of the true density $P_0$ with large probability. We denote the complement of the Hellinger neighborhood of radius $\epsilon_n$ by
 $$\mathcal{H}_{\epsilon_n}^c=\{\btheta:d_{\rm H}(P_0,P_{\btheta})> \epsilon_n\}.$$
Assume the following conditions:
\begin{enumerate}
    \item[A.1] The activation function $\psi(.)$ is 1-Lipschitz continuous upto a constant (e.g. relu, silu).
    \item[A.2] The hyperparameter $\sigma^2_0$ is fixed at a known constant.
    \item[A.3] The hyperparameters $a_0$, $b_0$ and $d_0$ are fixed constants.
\end{enumerate}

Assumption A.1 is a standard assumption on activation functions (see also assumption A.3, assumption 4.2 in \cite{Bai-Guang-2020}) and used to compute the covering number of the sieve $\mathcal{F}(L,\boldk,\bolds,\boldB)$. For simplicity, assumption A.2 assumes that the prior variance $\sigma_0$ is fixed (see Condition 4.3 in \cite{Bai-Guang-2020} and Corollary 4.5 in \cite{Jantre-et-al-2023}). For simplicity, Assumption A.3 assumes that the hyperparameters for the global scale parameters in \eqref{e:group-lasso-prior} and \eqref{e:group-horse-shoe-prior} are fixed constants (see Section \ref{sec:expts} for their choice).

\subsection{Posterior Consistency of spike-and-slab group lasso (SS-GL) and spike-and-slab group horseshoe (SS-GHS)}
Consider the prior in \eqref{e:group-lasso-prior}. Since $\varsigma^2$ follows $G(a_0,b_0)$, $1/\varsigma^2 \sim IG(a_0,b_0)$.  Let $\pi(t)$ be the density function of $T$. Assumption A.4  below quantifies the heavy tailed nature of Inverse Gamma distribution by assuming it behaves like Pareto with tail index $a''$ and an associated slowly varying function $L(t)$ which is identical to a constant as $t \to \infty$ \cite{beirlant2004statistics, horse-shoe-multi}.
\begin{enumerate}
    \item[A.4] If $T \sim IG(a_0,b_0)$, then $\pi(t)=K t^{-a''-1}L(t)$ for some constant $a''>0$ and $L(t)\geq c_0''$ for all $t\geq t_0''$ and some constant $c_0''>0$.
\end{enumerate}
For the posterior contraction rate of the variational posterior of SS-GL, consider the variational family in \eqref{e:group-lasso-var}. Define
\begin{equation}
\left.
\begin{aligned}
\label{e:gl-r}
\hspace{-2mm} u_l &= \log n+\log L+\sum  \log k_l+\sum \log k_{l+1}\\
\hspace{-2mm} \vartheta_l &= -\log \Big(\frac{B_l^2 }{(k_l\!+\!1)}\Big(\frac{1}{t_0''(k_l\!+\!1)}\Big)\Big)+ \frac{B_l^2 }{(k_l\!+\!1)}\Big(\frac{1}{t_0''(k_l\!+\!1)}\Big) \\
\hspace{-2mm} & \quad + 2\log n +2 L+2\sum \log B_m \\
\hspace{-2mm} r_l &= s_l (k_l\!+\!1)  \vartheta_l/n
\end{aligned}
\right\}
\end{equation}

\begin{theorem}
\label{thm:var-post}
Let A.1-A.4 hold. Let $\epsilon_n=\sqrt{(\sum_{l=0}^L r_l+\xi)\sum_{l=0}^L u_l}$ with  $r_l$, $u_l$ as in \eqref{e:gl-r} and $\xi$ as in \eqref{e:xi-def}.  For a sequence $M_n \to \infty$,  $M_n \epsilon_n \to 0$, the variational posterior $\widetilde{\Pi}^*$  satisfies, 
$$\widetilde{\Pi}^*(\mathcal{H}_{M_n\epsilon_n}^c) \stackrel{P_0^n}{\to} 0, \quad n \to \infty.$$
\end{theorem}

Theorem \ref{thm:var-post} quantifies the rate of contraction of variational posterior for SS-GL. The two main quantities that control the rate are $B_l^2/(k_l+1)$ and $1/(t_0''(k_l+1))$. Whereas $B_l^2/(k_l+1)$ is the effective strength of weights incident onto a node, $1/(t_0''(k_l+1))$ acts like penalty imposed on this effective strength. Lower values of $1/(t_0''(k_l+1))$ give better contraction rates but too small values may explode $-\log (B_l^2 /(t_0''(k_l+1)^2))$.

The following corollary gives conditions on hyperparameter  $\lambda_l$ for convergence of variational posterior under SS-GL.
\begin{corollary}
\label{cor:var-post}
For $\vartheta_l$ as in \eqref{e:gl-r} and $-\log \lambda_l=\log (k_{l+1})+C_l(k_l+1) \vartheta_l$, the conditions of Theorem \ref{thm:var-post} hold. 
\end{corollary}

For posterior contraction rate of the variational posterior of SS-GHS, consider the half-Cauchy prior in  \eqref{e:group-horse-shoe-prior}. As in \cite{horse-shoe-multi}, we make the following regularity assumption on half-Cauchy prior. Let $\pi(t)$ be the density function of random variable $T$.
\begin{enumerate}
    \item [A.5] If $\sqrt{T} \sim C^{+}(0,1)$, then $\pi(t)=Kt^{-a-1}L(t)$ for some constant $a>0$ and $L(t)\geq c_0$ for all $t\geq t_0$ and some constant $c_0>0$.
    \item [A.6] If $\sqrt{T} \sim C^{+}(0,s_0)$, then $\pi(t)=Kt^{-a'-1}L(t)$ for some constant $a'>0$ and $L(t)\geq c_0'$ for all $t\geq t_0'$ and some constant $c_0'>0$.
\end{enumerate}

Assumption A.5-A.6 quantifies the heavy tailed nature of Half Cauchy distribution (similar to Condition C1 in \cite{horse-shoe-multi}). Consider the prior formulation and variational family as in \eqref{e:group-horse-shoe-prior-split} and \eqref{e:group-horse-shoe-var} respectively. Define
\begin{equation}
\left.
\begin{aligned}
\label{e:ghs-r}
\hspace{-2mm} u_l&= \log n+\log L+\sum \log k_l+\sum \log k_{l+1}+\log c^2_{\rm reg}\\
\hspace{-2mm} \vartheta_l &=-\log\Big( \frac{B_l^2 }{(k_l\!+\!1)}\Big(\frac{1}{t_0t_0'}+\frac{1}{c^2_{\rm reg}}\Big)\Big)+ \frac{ B_l^2}{(k_l\!+\!1)}\Big(\frac{1}{t_0t_0'}+\frac{1}{c^2_{\rm reg}}\Big) \\
\hspace{-2mm} & \quad + 2\log n +2 L+2\sum \log B_m\\
\hspace{-2mm} r_l &= s_l (k_l\!+\!1)  \vartheta_l/n
\end{aligned}
\right\}
\end{equation}

\begin{theorem}
\label{thm:var-post-ssghs} Let
A.1-A.3, A.5-A.6 hold. With  $r_l$, $u_l$ in \eqref{e:ghs-r} and $\xi$ as in \eqref{e:xi-def}, let $\epsilon_n=\sqrt{(\sum_{l=0}^L r_l+\xi)\sum_{l=0}^L u_l}$. For a sequence $M_n \to \infty$,  $M_n \epsilon_n \to 0$, the variational 
 posterior $\widetilde{\Pi}^*$  satisfies, 
$$ \widetilde{\Pi}^*(\mathcal{H}_{M_n\epsilon_n}^c)\stackrel{P_0^n}{\to}0, \quad n \to \infty.$$
\end{theorem}

Theorem \ref{thm:var-post-ssghs} quantifies the contraction rate of variational posterior for SS-GHS. As in SS-GL, the main quantities are $B_l^2/(k_l+1)$ which is the effective strength incident onto a node and $(1/t_0t_0'+1/c^2_{\rm reg})$ which is the penalty imposed on this effective strength. As noted in Section \ref{sec:ssghs-prior}, $c_{\rm reg} \to \infty$ leads to unregularized version of SS-GHS. Indeed the results of Theorem \ref{thm:var-post-ssghs} continue to hold even $c^2_{\rm reg}$ is polynomial in $k_l$ (see also Appendix C of the supplement which demonstrate the similar performance of SS-GHS for $c^2_{\rm reg}=1$ and  $c^2_{\rm reg}\sim k_l$)

%Lower values of $(1/t_0t_0'+1/c^2_{\rm reg})$ lead to better shrinkage but too small values may cause $-\log (B_l(1/t_0t_0'+1/c^2_{\rm reg})/ (k_l+1))$ to explode. 

The following corollary gives conditions on hyperparameter  $\lambda_l$ for convergence of variational posterior under SS-GHS.

\begin{corollary}
\label{cor:var-post-ssghs}
For $\vartheta_l$ as in \eqref{e:ghs-r} and  $-\log \lambda_l=\log (k_{l+1})+C_l(k_l+1) \vartheta_l$, the conditions of Theorem \ref{thm:var-post-ssghs} hold. 
\end{corollary}

The contraction rates of both  SS-GL and SS-GHS depend on $r_l$ and $\xi$ where $r_l$ controls the number of nodes. If the NN is not sparse, $r_l$ is $k_{l+1}(k_l+1)\vartheta_l/n$ instead of $s_l (k_l+1)\vartheta_l/n $ which makes the convergence of $\epsilon_n\to 0$ difficult. If $s_l$ and $B_l^2/(k_l+1)$ are too small, $\xi$ may explode since a good approximation to the true function may not exist in a very sparse space.

\subsection{Discussion of theoretical derivations}
Similar to the previous work \cite{Jantre-et-al-2023}, the main challenge is to establish the rates $u_l$ and $\vartheta_l$ but with spike and slab Gaussian prior (SS-IG) replaced by SS-GL and SS-GHS  (see Appendix A for SS-GL and Appendix B for SS-GHS  in Supplement). To  compute $u_l$ (see Lemma \ref{lem:prior} in Appendix \ref{AppendixB}) for SS-GL, we exploited the tail behavior  and expectation under Gamma distribution. Since regularized horseshoe behaves like Gaussian in the tails, the proof of Lemma \ref{lem:prior} is similar to that in \cite{Jantre-et-al-2023}. The main challenge was to compute  $\vartheta_l$ (see Lemma \ref{lem:kl} in Appendix \ref{AppendixB})  for SS-GL and SS-GHS. For proof of Lemma \ref{lem:kl}(1), we exploit the heavy tailed nature of Inverse Gamma for SS-GL and Half-Cauchy for SS-GHS as highlighted in  Assumption A.4-A.6. For proof of Lemma \ref{lem:kl}(2) we carefully bound KL-divergence between Gamma and Log-normal distributions for SS-GL and Half Cauchy (reparameterized as product of Gamma and inverse Gamma) and Log-normal distribution for SS-GHS while ensuring  rates in Lemma \ref{lem:kl}(1) still hold. The following remark ties together the contraction rates of SS-GL, SS-GHS, and SS-IG under one big umbrella. It also provides theoretical insights into why one would expect the SS-GHS and SS-GL to outperform the SS-IG, a result observed across all experiments in Section \ref{sec:expts}.

\begin{remark}
Consider the generic form  $\vartheta_l = 2\log n + 2L -\log(\lambda_{\rm pen}B_l^2/(k_l+1)) + \lambda_{\rm pen}B_l^2/(k_l+1) + 2\sum \log B_m$ where $\lambda_{\rm pen}$ behaves like a penalization constant. For a simple spike-and-slab Gaussian prior and spike-and-slab mean field Gaussian variational family,  $\lambda_{\rm pen}=1$. As in Theorems \ref{thm:var-post} and \ref{thm:var-post-ssghs}, $\lambda_{\rm pen}$ is $1/(t_0''(k_l+1))$ and $(1/t_0t_0'+1/c^2_{\rm reg})$ for SS-GL and SS-GHS respectively.  Note, if for a nonparametric function, the best approximating neural network has exploding values of $B_l^2/(k_l+1)$, SS-GL with $t_0''(k_l+1) >1$ and SS-GHS with $1/t_0t_0''+1/c^2_{\rm reg}<1$ is expected to produce better contraction rates than Gaussian. On the other hand, if the best approximating neural network has shrinking values of $B_l^2/(k_l+1)$, SS-GL with $t_0''(k_l+1) <1$ and SS-GHS with $1/t_0t_0''+1/c^2_{\rm reg}>1$ is expected to produce better contraction rates than Gaussian. 
\end{remark}

Let $\mathcal{C}^{\alpha}_p$ be the class of $\alpha$-H$\ddot{\rm o}$lder smooth functions (defined on page 1880 of \cite{Schmidt-Hieber-2017}) on the unit cube $[0,1]^p$. For this class,  $\epsilon_n=n^{-\alpha/(\alpha+p)}$ is the optimal contraction rate (see \cite{Schmidt-Hieber-2017,Polson-Rockova-2018}) for edge sparse NNs with number of effective parameters $s$.
 
\begin{remark}
\label{rem:optimal-cont}
Let $L \sim \log n$, $k_l\sim n^{ p(1-\rho)/(2\alpha+p)}/\log n$, $\rho<1$ and $\eta_0 \in \mathcal{C}^{\alpha}_p$. Suppose there exists a network in $\mathcal{F}(L,\boldk,\bolds,\boldB)$ with $s_l=O( n^{ p\rho/(2\alpha+p)})$, $B_l^2=O( n^{p(1-\rho)/(2\alpha+p)}/\log n)$, $\xi=n^{-2\alpha/(2\alpha+p)}$, then the variational posterior for both SS-GHS and SS-GL contract at least at the rate $\epsilon_n=n^{-\alpha/(\alpha+p)}$.    
\end{remark}
For SS-GL and SS-GHS, the proof of Remark \ref{rem:optimal-cont} follows since the effective number of paramters is $\sum_{l=0}^L s_l\max( k_l, B_l^2)$ upto $\log n $ order.

A global sparsity of $s$ tries to find networks with no more than $s$ connections. However, a layer wise sparsity of $s_l$ tries to find networks with no more than $s_l$ nodes in each layer while allowing the connection strength to be $B_l^2$. Whereas 
 \cite{Schmidt-Hieber-2017} relies on the existence of networks with large number of nodes but sparsely connected, our approach relies on existence of networks with small number of nodes but almost fully connected. However, when networks are fully connected, the connection strength $B_l^2$ increases and thereby the need of shrinkage priors to penalize this strength with the penalty $\lambda_{\rm pen}$. 

%----------------------------------------------------------------------
\section{Experiments}
\label{sec:expts}
%----------------------------------------------------------------------

In this section, we demonstrate the performance of our proposed \textbf{SS-GL} and \textbf{SS-GHS} on network architectures and techniques used in practice. We consider multilayer perceptron (MLP), LeNet-5-Caffe, and ResNet architectures which we implement in PyTorch \cite{PyTorch2019NeurIPS}. We perform the image classification on  MNIST, Fashion-MNIST, and CIFAR-10 datasets.

In MLP and LeNet-5-Caffe experiments, we choose $\sigma^2_0=1$ after fine-tuning and observe minimal sensitivity to different choices. For our models, the layer-wise prior inclusion probabilities are defined by the formula: $\lambda_l = (1/k_{l+1})\exp(-C_l(k_l+1)\vartheta_l)$. The expression is similar for both SS-GL and SS-GHS, as noted in Corollaries~\ref{cor:var-post} and \ref{cor:var-post-ssghs}. We follow the guidance of \cite{Jantre-et-al-2023} for choosing $C_l$ values to be the negative order of 10 for ensuring the prior inclusion probabilities stay above $10^{-50}$. If otherwise $\lambda_l$ approaches 0, it eliminates all nodes in a given layer. Accordingly we chose $C_l = 10^{-9}$ in our experiments and provide further explanation in Appendix~\ref{appendix:constant_tuning}. Our numerical investigation supports the proposed choices of layer-wise prior inclusion probabilities in fully connected layers for both our models. This, in turn, underscores the significance of our theoretical developments. In SS-GL, we use $\varsigma^2 \sim G(a_0=4,b_0=2)$. However, model performance is not highly sensitive to different choices of $a_0$ and $b_0$. In SS-GHS, we choose $d_0^2=1$ and $c_{\rm reg}=1$. We further provide ablation experiment on regularization parameter (see Appendix C of the supplement), $c_{\rm reg}$, highlighting differences in results for regularized and unregularized versions of Horseshoe. The remaining tuning parameter details such as initial parameter choice, learning rate, and minibatch size are provided in Appendices~\ref{appendix:var_para_init} and \ref{appendix:hyper_train}. The prediction accuracy is calculated using a variational Bayes posterior mean estimator with 10 Monte Carlo samples at test time. We use swish (SiLU) activations \cite{Elfwing-et-al-2018, Ramachandran-et-al-2017} instead of ReLU for SS-IG as well as our SS-GL and SS-GHS models to avoid the dying neuron problem \cite{Lu-2020-DyingReLU}. Other smoother activation functions could be used but swish displayed the best performance.

\begin{figure*}[!ht]
\centering
\begin{subfigure}[b]{0.325\textwidth}
    \centering
\includegraphics[width=\textwidth]{Figs/MLP-MNIST/MLP_MNIST_Test_Accuracy_Overall.png} 
    \caption{{\small Prediction accuracy}}    
    \label{fig:MLP-MNIST-Overall-Test-Acc}
\end{subfigure}
\begin{subfigure}[b]{0.325\textwidth}
    \centering
\includegraphics[width=\textwidth]{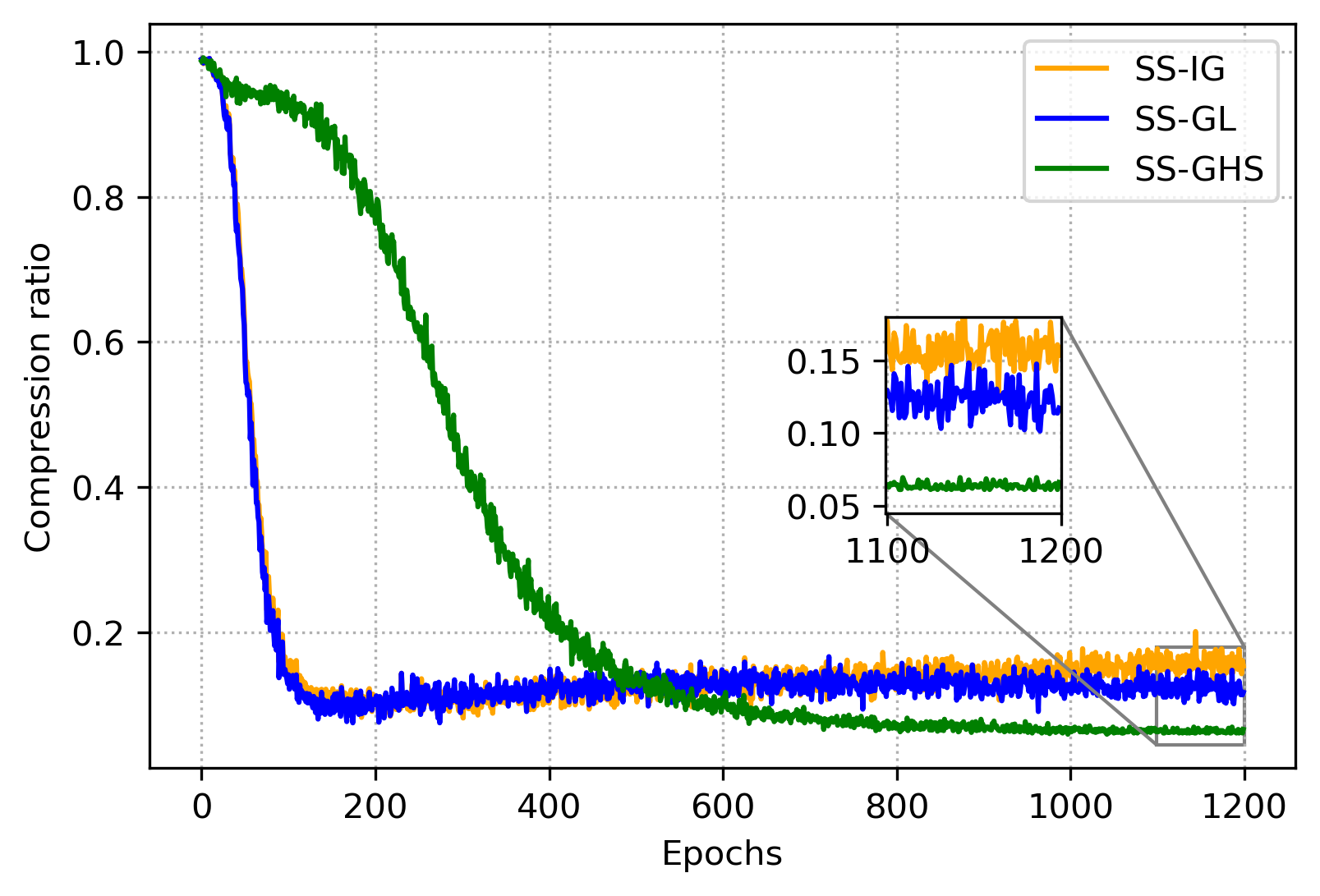} 
    \caption{{\small Compression ratio}}    
    \label{fig:MLP-MNIST-Overall-Compression-Ratio}
\end{subfigure}
\begin{subfigure}[b]{0.325\textwidth}
  \centering
\includegraphics[width=\linewidth]{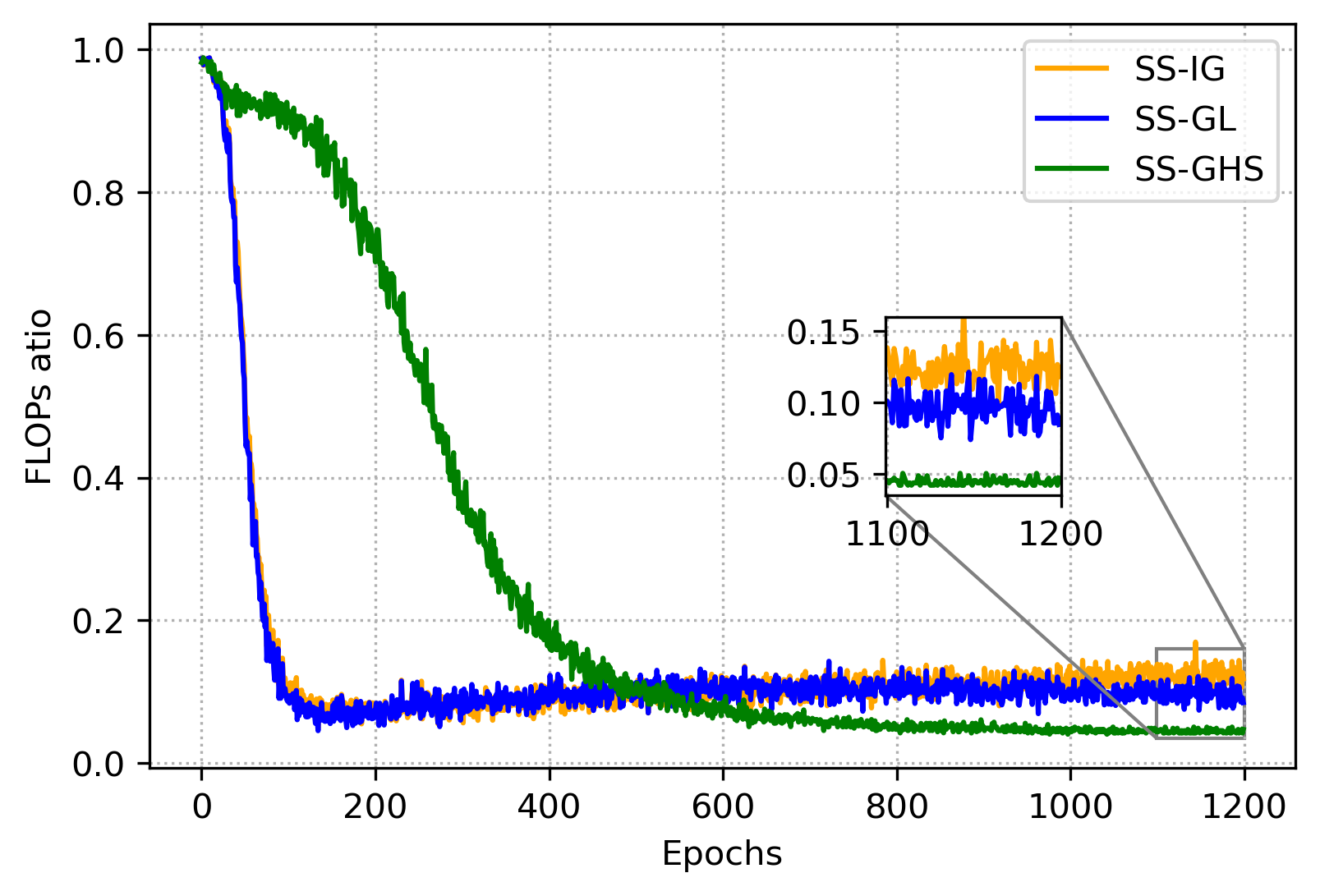}
  \caption{{\small FLOPs ratio}}
  \label{fig:MLP-MNIST-Overall-Flops-Ratio}
\end{subfigure}
\caption{{\bfseries MLP-MNIST experiment.} We compare the performance of SS-GHS, SS-GL, and SS-IG models on (a) classification accuracy, (b) model compression ratio, and (c) FLOPs ratio. The SS-GHS model achieves the highest accuracy, best compression, and lowest FLOPs, making it the most efficient in performance and resource use.}
\label{fig:MLP-MNIST-Overall-Experiments}
\end{figure*}

We provide node sparsity estimates for each linear hidden layer separately. In all models, for a fully connected linear layer, the layer-wise node sparsity is given by
\begin{equation}
\label{eq:node-sparsity}
{\rm Node \: sparsity} = \frac{O_{pr}}{O}     
\end{equation}
where $O_{pr}$ represents number of neurons with at least one nonzero incoming edge and $O$ represents total number of neurons present in that layer before training. We further extend our SS-GL and SS-GHS models to convolutional neural network (CNN) architectures for numerical investigation. For convolutional layers, we prune output channels--similar to pruning neurons in linear layers--using our spike-and-slab prior. Each output channel is assigned a Bernoulli variable to collectively prune its associated parameters. The layer-wise channel sparsity estimate is provided by:
\begin{equation}
\label{eq:channel-sparsity}
{\rm Channel \: sparsity} = \frac{C_{out,pr}}{C_{out}}
\end{equation}
where $C_{out,pr}$ represents number of output channels with at least one nonzero incoming connection and $C_{out}$ represents total number of output channels present in that layer before training. The layer-wise node or channel sparsity estimates provide a granular illustration of the structural compactness of the trained model. The structural sparsity in the trained model leads to lower computational complexity at test time which is vital for resource-constrained devices.

\vspace{0.1in}
\noindent {\bf Additional metrics.}
We provide two additional metrics that relate to the model compression and computational complexity. (i) {\it compression ratio:} it is the ratio of the number of nonzero weights in the compressed network versus the dense model and is an indicator of storage cost at test-time. (ii) {\it floating point operations (FLOPs) ratio:} it is the ratio of the number of FLOPs required to predict the output from the input during test-time in the compressed network versus its dense counterpart. We have detailed the FLOPs calculation in Appendix~\ref{appendix:flops}. Layer-wise node sparsities (\ref{eq:node-sparsity}) and channel sparsities (\ref{eq:channel-sparsity}) are directly related to FLOPs ratio and we only provide FLOPs ratio in LeNet-5-Caffe and ResNet models.

\subsection{MLP MNIST Classification}
\label{MLP-MNIST}
In this experiment, we use MLP model with 2 hidden layers having 400 nodes per layer to fit the MNIST data which consists of 60,000 small square 28×28 pixel grayscale images of handwritten single digits between 0 and 9. We preprocess the images in the MNIST data by dividing their pixel values by 126. The output layer has 10 neurons since there are 10 classes in the MNIST data. We compare our proposed models against closely related spike-and-slab prior based structured sparsity baseline -- SS-IG -- to demonstrate the improvements in model performance as well as reduction across all metrics. The results are presented in Fig.~\ref{fig:mnist-results-intro} and \ref{fig:MLP-MNIST-Overall-Experiments}. We provide prediction accuracy trajectory for test data over the training period (Fig.~\ref{fig:MLP-MNIST-Overall-Test-Acc}). We provide layer-wise node sparsities for both layers (Fig.~\ref{fig:MLP-MNIST-layer1-node-sparsity} and \ref{fig:MLP-MNIST-layer2-node-sparsity}), model compression ratio (Fig.~\ref{fig:MLP-MNIST-Overall-Compression-Ratio}), and FLOPs ratio (Fig.~\ref{fig:MLP-MNIST-Overall-Flops-Ratio}) learning trajectories to highlight the dynamic structural compactness of the models during training (Fig.~\ref{fig:MLP-MNIST-layer1-node-sparsity} and \ref{fig:MLP-MNIST-layer2-node-sparsity}).

\vspace{0.1in}
\noindent {\bf MLP-MNIST comparison with SS-IG:}
In Fig.~\ref{fig:MLP-MNIST-Overall-Test-Acc}, we observe that SS-GHS has better predictive accuracy compared to SS-GL and SS-IG models. In Fig.~\ref{fig:MLP-MNIST-layer1-node-sparsity} and \ref{fig:MLP-MNIST-layer2-node-sparsity}, SS-GHS prunes maximum number of nodes in both layers in contrast to SS-GL and SS-IG models. This in turn, leads to a compact model architecture for SS-GHS that requires minimal number of FLOPs during test-time inference, demonstrating its low-latency inference capability (Fig.~\ref{fig:MLP-MNIST-Overall-Flops-Ratio}). Moreover, SS-GHS leads to a compact model requiring minimal storage among the node selection models compared (Fig.~\ref{fig:MLP-MNIST-Overall-Compression-Ratio}). In the model compression plots, SS-GHS takes longer to reach a compact structure, likely due to heavy tails of half-Cauchy distribution  (\ref{e:group-horse-shoe-prior}) in contrast to the light-tails of Gamma distribution in SS-GL (\ref{e:group-lasso-prior}) and Gaussian distribution in SS-IG. Lastly, SS-GL and SS-IG have similar predictive accuracies; however, SS-GL has lower layer-wise node sparsities in both layers, hence lower FLOPs, and it also has storage cost compared to SS-IG.
\vspace{-0.1in}

\subsection{LeNet-5-Caffe Experiments}
\label{Lenet-Caffe-MNIST}
We evaluate the compression ability of SS-GL and SS-GHS models in CNN architectures using LeNet-5-Caffe network on MNIST and Fashion-MNIST \cite{xiao2017fashionmnist} datasets. In Section~\ref{sec:theory}, we provide variational posterior consistency for SS-GL and SS-GHS under the MLP network assumption. This can be extended to CNNs (see Section 3.4.1 in \cite{Gal-2016}), as each convolutional operation can be seen as a linear mapping with a Toeplitz weight matrix. We leave this theoretical development for future work. To this end, our CNN experiments aim to demonstrate the efficacy of our proposed models beyond the MLP experiments. The LeNet-5-Caffe model has 2 convolutional layers (20 and 50 feature maps) with 5x5 filters, followed by 2x2 max pooling. The flattened features (size=800) feed into 2 hidden layers (800 and 500 neurons) and an output layer with 10 neurons. For comparison, we include dense baselines of fully connected deterministic (DNN) as well as mean field VI inferred Bayesian neural networks (BNN). Moreover, we include node-sparse methods:  node-sparse random dropout \cite{srivastava2014dropout} (NS-Drop), NS-MC node-sparse MC-dropout (NS-MC), Bayesian compression using group normal-Jeffreys prior (BC-GNJ), and SS-IG. Detailed description of the baseline models compared are provided in Appendix~\ref{appendix:additional-models} and their fine tuned hyperparameters used for training are given in Appendix~\ref{appendix:hyper_train}.

In Table~\ref{Table:LeNet_Expts}, we report the means and standard deviations of the performance measures from three independent runs. We quantify the predictive performance using the accuracy of the test data. Besides the test accuracy, we use {\it compression (\%)} and {\it pruned FLOPs (\%)} which are compression ratio and FLOPs ratio discussed earlier converted to percentages respectively. The results show that, SS-GHS consistently achieves the lowest inferential cost (pruned FLOPs) in both MNIST and Fashion-MNIST datasets. Furthermore, the predictive performance (test accuracy) of SS-GHS is close to dense BNN highlighting its effectiveness in maintaining model performance while achieving lowest test-time latency. The model storage costs are also lowest for SS-GHS in Fashion-MNIST case and close to BC-GNJ in MNIST case. Lastly, SS-GL model consistently outperforms SS-IG model and achieves second best inferential cost in Fashion-MNIST case.

\begin{table}[t!]
% \fontsize{7.4}{9}\selectfont
\centering
    \begin{tabular}{|c|l|crr|}
        \hline
        Data & Method & Test Accuracy & Compression (\%) & Pruned FLOPs (\%)\\
        \hline
        M & DNN & 99.31 (0.08) & 100.00 (0.00) & 100.00 (0.00) \\ 
        & BNN & 99.03 (0.03) & 100.00 (0.00) & 100.00 (0.00) \\
        & NS-Drop & 98.87 (0.11) & 16.43 (0.00) & 18.52 (0.00) \\
        & NS-MC & 98.91 (0.08) & 14.03 (1.90) & 16.43 (3.60) \\
        & BC-GNJ & \textbf{99.21 (0.08)} & 2.70 (0.08) & 13.26 (2.29) \\
        & SS-IG  & 98.68 (0.09) & 6.22 (2.70) & 15.05 (0.77) \\
        & SS-GL (ours)  & 98.88 (0.14) & 4.25 (1.24) & 14.25 (1.66) \\
        & SS-GHS (ours) & 98.98 (0.09) & 3.01 (0.16) & \textbf{12.55 (0.83)} \\
        \hline
        FM & DNN & 91.48 (0.19) & 100.00 (0.00) & 100.00 (0.00) \\ 
        & BNN & 90.45 (0.03) & 100.00 (0.00) & 100.00 (0.00) \\
        & NS-Drop & 89.39 (0.17) & 16.43 (0.00) & 18.52 (0.00) \\
        & NS-MC & 89.36 (0.17) & 14.49 (1.26) & 16.95 (4.38) \\
        & BC-GNJ & \textbf{90.88 (0.24)} & 3.82 (0.18) & 25.13 (1.34) \\
        & SS-IG & 87.22 (0.30) & 6.47 (0.48) & 14.22 (0.80) \\
        & SS-GL (ours)  & 88.32 (0.04) & 5.22 (0.98) & 13.71 (0.91) \\
        & SS-GHS (ours) & 89.50 (0.32) & 3.37 (0.10) & \textbf{13.36 (1.67)} \\ 
        \hline
    \end{tabular}
    \caption{{\bf LeNet-5-Caffe experiments.} M and FM are MNIST and Fashion-MNIST datasets, respectively. The results of LeNet-5-Caffe on each dataset are calculated by averaging over 3 independent runs with standard deviation reported in parentheses. We highlight the performance of best metrics among sparse models in bold.}
    \label{Table:LeNet_Expts}
\end{table}

\subsection{Residual Network Experiments}
\label{ResNet-CIFAR10}
This section presents experiments using Residual Networks (ResNet) \cite{He-ResNet-2016} applied on CIFAR-10 \cite{He-ResNet-2016}. The dataset consists of 60000 32x32 color images in 10 classes. There are 50000 training images and 10000 test images. We employ ResNet-20 and ResNet-32 architectures, training them for 300 epochs with SGD using momentum $(=\!0.9)$ and a mini-batch size of 128. Our training procedure includes batch normalization, a stepwise (piecewise constant) learning rate schedule where we divide the learning rate by 10 at epochs 150 and 225, and augmented training data \cite{Random_Erasing_2020}. For comparison, we include dense baselines-- DNN and BNN, node sparse methods-- NS-Drop, NS-MC, BC-GNJ, and SS-IG, edge sparse methods-- consistent sparse deep learning (BNN$_{\rm cs}$) \cite{Sun-Liang-2021}, and variational Bayes neural network with mixture Gaussian prior (VBNN) \cite{blundell2015weight}. We provide unstructured sparsity methods to demonstrate the trade-off between the inferential cost and memory cost at test-time among our structured sparsity methods and recent unstructured sparsity methods. Details of the models compared are provided in Appendix~\ref{appendix:additional-models} and their fine tuned hyperparameters used for training are given in Appendix~\ref{appendix:hyper_train}.

In this experiment, $\sigma^2_0=0.04$ is chosen after careful fine-tuning. In SS-IG, SS-GL, and SS-GHS, we use common prior inclusion probabilities in each convolution layer, $\lambda_l = 10^{-4}$, after hyperparameter search since our theory does not cover Bayesian CNNs. In SS-GL, we use $\varsigma^2 \sim G(a_0=4,b_0=2)$ with the model performance being less sensitive to different $a_0$ and $b_0$ choices. In SS-GHS, we choose $d_0^2=1$ and $c_{\rm reg}=1$. We quantify the predictive performance using the test data accuracy, inferential cost using pruned FLOPs (\%), and storage cost using compression (\%). In this experiment, we only count parameters and FLOPs over the convolution and last fully connected layer, because our proposed methods focus on channel and node pruning of convolution and linear layers respectively. For ResNet, our proposed methods under centered parameterization as in  previous experiments of Section~\ref{MLP-MNIST} and \ref{Lenet-Caffe-MNIST} have unstable performance. Instead, we use non-centered parameterization \cite{Ghosh-JMLR-2018} to stabilize the training.

\begin{table}[!t]
% \fontsize{7.2}{9}\selectfont
\centering
    \begin{tabular}{|c|l|crr|}
        \hline
        Model & Method & Test Accuracy & Compression (\%) & Pruned FLOPs (\%)\\
       \hline
        RN-20 & DNN & 91.99 (0.06) & 100.00 (0.00) & 100.00 (0.00) \\
        & BNN & 92.96 (0.07) & 100.00 (0.00) & 100.00 (0.00) \\
        & NS-Drop & 91.42 (0.16) & 98.82 (0.00) & 99.50 (0.00) \\
        & BNN$_{\rm cs}$ (20\%)  & 92.23 (0.16) & 19.29 (0.12) & 98.94 (0.38) \\
        & BNN$_{\rm cs}$ (10\%)  & 91.43 (0.11) & 9.18 (0.13) & 99.13 (0.37) \\
        & VBNN (20\%)  & 89.61 (0.04) & 19.55 (0.01) & 100.00 (0.00) \\
        & VBNN (10\%)  & 88.43 (0.13) & 9.50 (0.00) &  99.93 (0.00) \\
        & NS-MC & 90.43 (0.35) & 89.15 (0.68) & 88.45 (0.28) \\
        & BC-GNJ & 92.27 (0.14) & 98.35 (0.25) & 99.19 (0.15) \\
        & SS-IG & 92.94 (0.15) & 79.52 (0.98) & 88.39 (1.00) \\
        & SS-GL (ours)  & \textbf{92.99 (0.11)} & 76.10 (1.55) & \textbf{85.15 (1.76)} \\
        & SS-GHS (ours) & 92.87 (0.23) & 78.70 (0.42) & 86.18 (1.02) \\
        \hline
        RN-32 & DNN & 93.03 (0.09) & 100.00 (0.00) & 100.00 (0.00) \\
        & BNN & 93.56 (0.18) & 100.00 (0.00) & 100.00 (0.00) \\
        & NS-Drop & 92.04 (0.20) & 98.82 (0.00) & 99.49 (0.00) \\
        & BNN$_{\rm cs}$ (10\%)  & 92.65 (0.03) & 9.15 (0.03) &  94.53 (0.86) \\
        & BNN$_{\rm cs}$ (5\%)  & 91.39 (0.08) & 4.49 (0.02) & 90.79 (1.35) \\
        & VBNN (10\%) & 89.37 (0.04) & 9.61 (0.01) & 99.99 (0.02) \\
        & VBNN (5\%) & 87.38 (0.22) & 4.59 (0.01) & 94.27 (0.54)  \\
        & NS-MC & 90.65 (0.24) & 76.14 (0.18) & 75.43 (0.10)\\
        & BC-GNJ & 92.67 (0.40) & 87.27 (2.57) & 90.94 (0.82) \\
        & SS-IG  & 93.08 (0.23) & 55.28 (2.96) & 67.59 (2.36) \\
        & SS-GL (ours)  & \textbf{93.33 (0.11)} & 54.27 (1.73) & 66.93 (2.98) \\
        & SS-GHS (ours) & 93.15 (0.23) & 53.72 (2.11) & \textbf{66.68 (2.75)} \\
        \hline
    \end{tabular}
    \caption{{\bf ResNet-CIFAR-10 experiments.} RN-20 and RN-32 are ResNet-20 and ResNet-32 models respectively. The results of each method are calculated by averaging over 3 independent runs with standard deviation reported in parentheses. For BNN$_{\rm cs}$ and VBNN, we show predefined percentages of pruned parameters used for magnitude pruning as in \cite{Sun-Liang-2021}.}
    \label{Table:ResNet_Expts}
\end{table}

\vspace{0.1in}
\noindent {\bfseries Non-Centered Parameterization:} We adopt non-centered parameterization for the slab component in both prior setups to circumvent pathological funnel-shaped geometries associated with coupled posterior \cite{Ingraham-Marks-2017, Ghosh-JMLR-2018}. Let $\tau_{lj}^{*2} = \tau_{lj}^{2}$ (for SS-GL) or $\tau_{lj}^{*2} = \widetilde{\tau}_{lj}^{2} s^2$ (for SS-GHS) leads to the following reparameterization of the priors in \eqref{e:group-lasso-prior} and \eqref{e:group-horse-shoe-prior-split}
$$\pi({\boldw}_{lj}|z_{lj}) =(1-z_{lj})\bdelta_0 + z_{lj} \tau_{lj}^*{\boldw}_{lj}^{(0)} , \hspace{3mm} \pi({\boldw}_{lj}^{(0)} )= N(0,\sigma_0^2 \boldI)$$
Variational families in \eqref{e:group-lasso-var} and \eqref{e:group-horse-shoe-var} are reparameterized as
\begin{align*}
q({\boldw}_{lj}|z_{lj}) &= (1-z_{lj})\bdelta_0+ z_{lj} \tau_{lj}^*{\boldw}_{lj}^{(0)} \\
q({\boldw}_{lj}^{(0)}) &= N(\bmu_{lj},\text{diag}(\bsigma^2_{lj}))
\end{align*}
This formulation leads to simpler posterior geometries \cite{Betancourt-Girolami-2015}. The non-centered reparameterization leads to efficient posterior inference without change in the form of the prior but with a minor change in the form of the variational family.

We summarize the ResNet experiment results using mean and standard deviations of performance metrics over three independent runs in Table~\ref{Table:ResNet_Expts}. We observe that, our proposed models outperform all the models in reducing inferential cost-- SS-GL in ResNet-20 and SS-GHS in ResNet-32 experiment. Moreover, we maintain the predictive performance similar to dense BNN model while outperforming all the sparsity inducing methods. Comparing with BNN${\rm cs}$ and VBNN models, our models demonstrate superior performance with significantly fewer FLOPs during inference at test time. While unstructured sparsity models like BNN${\rm cs}$ and VBNN have predefined high levels of pruned parameters, our structured sparsity models strike a balance, offering a reduction in storage cost and computational complexity at test time. This emphasizes the trade-off between unstructured and structured sparsity methods, and showcases the advantages of our approach. This overall experiment demonstrates the advantage of using spike-and-slab based group shrinkage priors for achieving high predictive performance with lower inferential cost and memory footprint.

\section{Conclusion}
In this paper, we have used spike and slab group lasso (SS-GL) and spike and slab group horseshoe (SS-GHS) to facilitate optimal node recovery. Our theoretical developments (Section \ref{sec:theory}) give  the choice hyperparameters for SS-GL and SS-GHS to guarantee the  contraction of the corresponding variational posterior. By exploiting the mathematical properties of heavy tailed distributions, we discuss the theoretical comparison of SS-GL and SS-GHS with the simple spike and slab Gaussian (SS-IG). As a final contribution, in the context of node selection (Section \ref{sec:expts}), we demonstrate the numerical efficacy of using SS-GL and SS-GHS. For a wide range of experiments including LeNet-5-Caffe and Residual Networks, we demonstrate that the node selection using SS-GL and SS-GHS lead to the smallest FLOPs ratio among all baselines. 

\section{Limitations and Future Work}
To further consolidate the mathematical findings in Section \ref{sec:theory}, it will be interesting to quantify how well a generic non parametric function can be approximated by  almost fully connected architectures but with small number of nodes. This will require a detailed study of topological spaces of nonparametric function in conjunction with the neural networks in the future. In Section \ref{sec:expts}, we observe a slight improvement in predictive performance and deterioration in convergence rate of SS-GHS over SS-GL. Providing a statistical validation to this phenomenon can be explored in future. Given the promising numerical performance of SS-GHS and SS-GL for LeNet-5-Caffe and Residual Networks, an immediate extension can be to generalize the theory developed for MLPs to CNNs.

\appendix
%----------------------------------------------------------------------
\section{Additional numerical experiments details}
\label{AppendixE}
%----------------------------------------------------------------------
\subsection{FLOPs Calculation}
\label{appendix:flops}
We only count multiply operations for floating point operations similar to \cite{Zhao2019_VarCNNPrune} in ResNet experiments. In 2D convolution layer, we assume convolution is implemented as a sliding window and the activation function is computed for free. Then, for a 2D convolution layer (given bias) we have
$$ {\rm FLOPs} = (C_{in,pr} K_w K_h + 1) O_w O_h C_{out,pr} $$
where $C_{in,pr}$ and $C_{out,pr}$ denote number of input and output channels after pruning, respectively. $C_{in,pr}$ for layer $l+1$ is equal to $C_{out,pr}$ for layer $l$ and for the first layer it is equal to $C_{in}$ since inputs are not pruned. Channels are pruned if all the parameters associated with that channel are zero. $K_w$ and $K_h$ are kernel width and height respectively. $O_w, O_h$ are output width and height where $O_w = (I_w + 2\times P_w - D_w \times (K_w - 1) - 1)/S_w+1$ and $O_h = (I_h + 2\times P_h - D_h \times (K_h - 1) - 1)/S_h+1$. Here, $I_w, I_h$ are input, $P_w,P_h$ are padding, $D_w,D_h$ are dilation, $S_w, S_h$ are stride widths and heights respectively. For fully connected (linear) layers (with bias) we have
$$ {\rm FLOPs} = (I_{pr}+1) O_{pr} $$ $I_{pr}$ and $O_{pr}$ are number of pruned input and output neurons. $I_{pr} = I$ and $O_{pr} = O$ for first layer and last layer, respectively, since we do not prune model inputs and outputs.

\subsection{Variational parameters initialization}
\label{appendix:var_para_init}
$\gamma_{lj}$s are initialized close to 1 ensuring a fully connected network at epoch 0. $\mu_{ljj'}$ are initialized using Kaiming uniform initialization \cite{Kaiming-He-2015}. We reparameterize $\sigma_{ljj'}$  with softplus function: $\sigma_{ljj'}=\log(1+\exp(\rho_{ljj'}))$ and set
$\rho_{ljj'}$ to -6. Thus, initial values of $\sigma_{ljj'}$ are close to 0 such that initial values of network weights stay close to Kaiming uniform initialization. 

In SS-GL, $\mu^{\{\tau\}}_{lj}$ is initialized with $U(-0.6,0.6)$ and ${\sigma^{\{\tau\}}_{lj}}=\log(1+\exp(\rho^{\{\tau\}}_{lj}))$ where $\rho^{\{\tau\}}_{lj}$ are initialized to -6. Further, $\mu_{\varsigma}$ is initialized to 1 and $\sigma_{\varsigma}=\log(1+\exp(\rho^{\{\tau\}}))$ where $\rho^{\{\tau\}}$ is initialized to -6. In SS-GHS, both $\mu^{\{\alpha\}}_{lj}$ and $\mu^{\{\beta\}}_{lj}$ are initialized with $U(-0.6,0.6)$. Also, ${\sigma^{\{\alpha\}}_{lj}}=\log(1+\exp(\rho^{\{\alpha\}}_{lj}))$ and ${\sigma^{\{\beta\}}_{lj}}=\log(1+\exp(\rho^{\{\beta\}}_{lj}))$ where $\rho^{\{\alpha\}}_{lj}$ and   $\rho^{\{\beta\}}_{lj}$ initialized to -6. 
Moreover, $\mu^{\{\zeta_a\}}$ and $\mu^{\{\zeta_b\}}$ are initialized to 1. Lastly $\sigma^{\{\zeta_a\}}=\log(1+\exp(\rho^{\{\zeta_a\}}))$ and $\sigma^{\{\zeta_b\}}=\log(1+\exp(\rho^{\{\zeta_b\}}))$ where $\rho^{\{\zeta_a\}}$, $\rho^{\{\zeta_b\}}$ are initialized to -6.

\subsection{Additional models for comparison}
\label{appendix:additional-models}
In Lenet-5-Caffe and ResNet experiments, we include deterministic neural network (DNN) representing frequentist dense baseline  Similarly, dense Bayesian neural network with Gaussian prior on network parameters inferred using mean field variational inference represents dense Bayesian baseline. For NS-Drop, we randomly prune output neurons in linear layers and output channel in convolution layers with dropout rate-- $p$. Similar to how MC-dropout \cite{gal2016dropout} uses weights obtained by Dropout \cite{srivastava2014dropout} multiplied by random masking vectors, we include node-sparse MC-Dropout (NS-MC) using node-wise sparse weights obtained by Targeted Dropout \cite{gomez2019targeteddropout} multiplied by random masking vectors. In particular, let $g_0 \in [0,1]$ and $h_0 \in [0,1]$ represent the pre-specified pruning and drop probabilities, respectively. Before each gradient step, we select the $g_0\times100\%$ of nodes with the lowest magnitude ($L_2$-norm of the connected weights) in each layer and randomly mask these nodes with probability $h_0$. This means the expected survival ratio of nodes during each step is $1-g_0h_0$. BC-GNJ adopts the Bayesian compression using group-normal-Jeffreys prior for pruning output neurons and channels from \cite{Louizos-et-al-2017}. In ResNet experiments, we additionally include unstructured pruning methods. First is consistent sparse deep learning (BNN$_{\rm cs}$) \cite{Sun-Liang-2021} which uses continuous relaxation of spike-and-slab Gaussian prior using mixture of Gaussians prior, where the spike component is replaced by a small variance Gaussian. Second is variational Bayesian neural network (VBNN) which uses mixture of Gaussians prior.

\subsection{Hyperparameters for training}
\label{appendix:hyper_train}
In MLP experiments, we use $10^{-3}$ learning rate and 1024 minibatch size for all three models -- SS-IG, SS-GL, and SS-GHS, and train them for 1200 epochs using Adam optimizer. In LeNet-5-Caffe experiments, we use $10^{-3}$ learning rate and 1024 minibatch size for BC-GNJ, SS-IG, SS-GL, and SS-GHS, and train them for 1200 epochs using Adam optimizer. For DNN, BNN, NS-Drop, and NS-MC we use $10^{-4}$ learning rate and 128 minibatch size and train them for 1200 epochs using Adam optimizer. Moreover, we chose dropout rate of $p=0.6$ for NS-Drop, $(g_0,h_0)=(0.8,0.8)$ for NS-MC, and threshold parameter for pruning is $-3$ for BC-GNJ after careful fine tuning. In ResNet experiments, all models are trained for 300 epochs with SGD using momentum $(=\!0.9)$ and a mini-batch size of 128. Our training procedure includes batch normalization, a stepwise (piecewise constant) learning rate schedule where we divide the learning rate by 10 at epochs 150 and 225, and augmented training data \cite{Random_Erasing_2020}. Lastly, we chose dropout rate of $p=0.02$ for NS-Drop, $(g_0,h_0)=(0.5,0.2)$ in ResNet-20 and  $(g_0,h_0)=(0.8,0.3)$ in ResNet-32 for NS-MC, and threshold parameter for pruning is $-3$ for BC-GNJ after careful fine tuning. In BNN$_{\rm cs}$ and VBNN, the prior hyperparameters are exactly same as reported by \cite{Sun-Liang-2021}.

\subsection{Fine tuning the constant in prior inclusion probabilities}
\label{appendix:constant_tuning}
In our SS-GL and SS-GHS models, the layer-wise prior inclusion probabilities are given by: 
$$\lambda_l=\frac{1}{k_{l+1}}\exp(-C_l (k_{l}+1)\vartheta_l)$$ 
as detailed in Corollaries~\ref{cor:var-post} and \ref{cor:var-post-ssghs}. The $\lambda_l$ value depends on the constant $C_l$, and we describe how to select it. $C_l$ primarily influences the inclusion probability through the factor $k_{l}+1$ and the term ${B_l}^2/(k_l+1)$ from $\vartheta_l$. To manage this, we ensure that the incoming weights and biases onto each node from layer $l+1$ is bounded by 1, leading us to set $B_l=k_l+1$. Consequently, the leading term in $(k_{l}+1)\vartheta_l$ simplifies to $(k_{l}+1)$. To avoid making the exponential term $\lambda_l$ expression close to zero, which would potentially prune all nodes in a layer and destabilize training, we choose $C_l$ values on the negative order of 10. This ensures that the prior inclusion probabilities $\lambda_l$ stay above $10^{-50}$. By carefully selecting $C_l$ in this manner, we maintain reasonable values for $\lambda_l$ across all layers, thereby ensuring stable and effective training of the network.

\section{SS-GHS regularization constant choice.} 
\label{Appendix-SS-GHS}
In Corollary 4, we provide the contraction rates for variational posterior in SS-GHS model for the regularization constant $c_{\rm reg}$. Here we experiment with two values of the constant $c_{reg}=1$ and $c_{\rm reg}=k_l+1$. The choice of $c_{reg}=1$ makes the Horseshoe tails closer to Gaussian whereas $c_{\rm reg}=k_l+1$ makes it closer to the unregularized Horseshoe. Here, we provide an MLP-MNIST experiment using SS-GHS model with aforementioned $c_{\rm reg}$ values. In MLP, the $k_l+1=400+1=401$ is a large constant and essentially acts as an unregularized model. We ran an unregularized version of the model and verified this claim but do not provide the results for brevity.

\begin{figure*}[!ht]
\centering
\begin{subfigure}[b]{0.325\textwidth}
    \centering
\includegraphics[width=\textwidth]{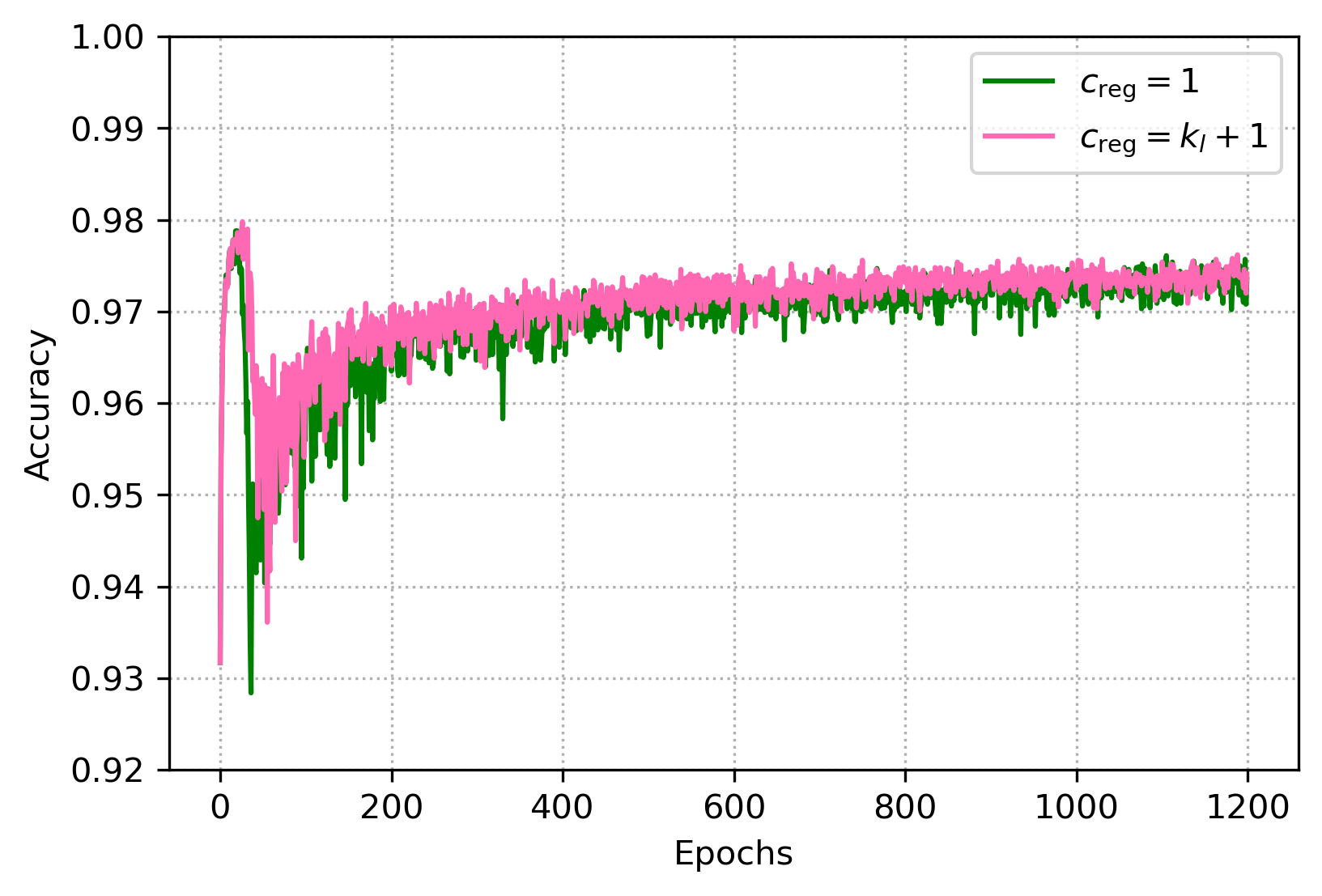} 
    \caption{{\small Prediction accuracy}}    
    \label{fig:MLP-MNIST-GHS-Test-Acc}
\end{subfigure}
\begin{subfigure}[b]{0.325\textwidth}
  \centering
\includegraphics[width=\linewidth]{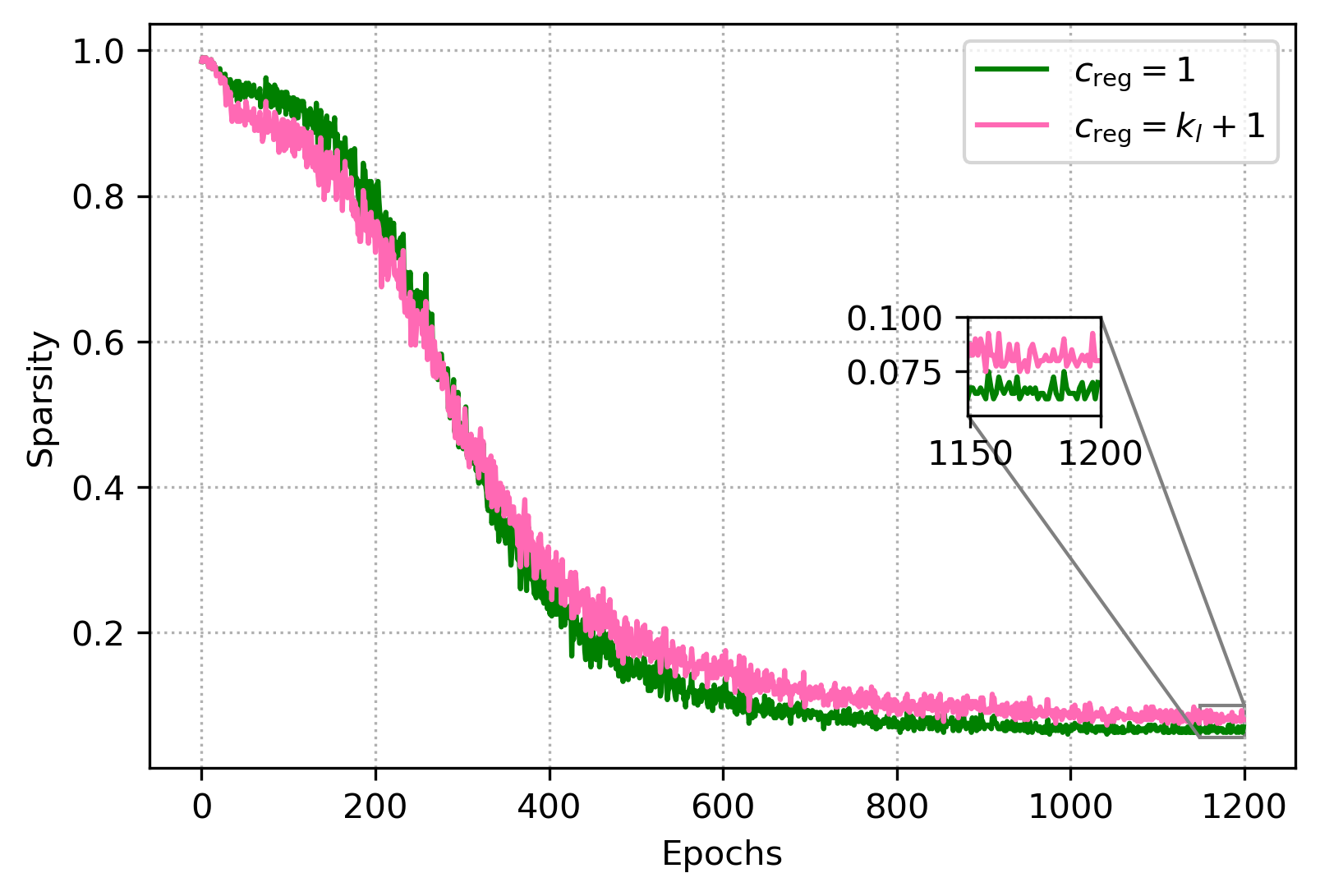}
  \caption{{\small Layer-1 node sparsity}}
  \label{fig:MLP-MNIST-GHS-layer1-sparsity}
\end{subfigure}
\begin{subfigure}[b]{.325\textwidth}
  \centering
  \includegraphics[width=\linewidth]{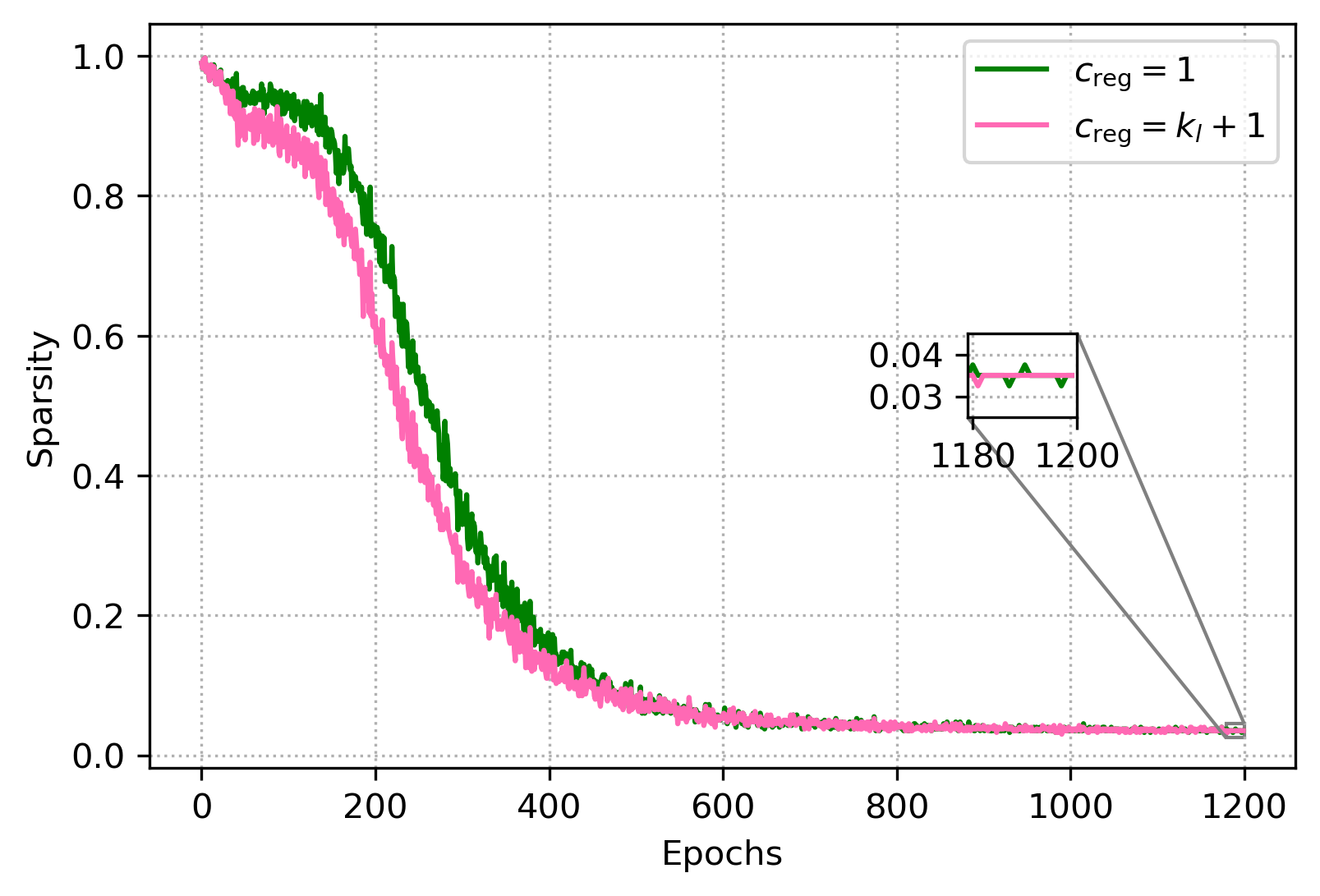}
  \caption{{\small Layer-2 node sparsity}}
  \label{fig:MLP-MNIST-GHS-layer2-sparsity}
\end{subfigure}
\caption{{\bf SS-GHS $c_{\rm reg}$ choice experiment.} Performance of SS-GHS with regularization constant of $c_{\rm reg}=1$  and $c_{\rm reg}=k_l+1=401$. (a) Classification accuracy on the test data. (b) and (c) proportion of active nodes (node sparsity) in layer-1 and layer-2 of the network respectively. Both $c_{\rm reg}$ choices give similar classification accuracies and $c_{\rm reg}=1$ has better layer-1 node sparsity.}
\label{fig:MLP-MNIST-GHS-Experiments}
\end{figure*}

Fig.~\ref{fig:MLP-MNIST-GHS-Experiments} summarizes the results of MLP-MNIST experiment using SS-GHS model $c_{\rm reg}=1$  and $c_{\rm reg}=k_l+1$. We observe that both values of $c_{\rm reg}$ lead to same predictive accuracies on test data. However in $c_{\rm reg}=1$ scenario, SS-GHS model has better layer-1 node sparsity (Fig.~\ref{fig:MLP-MNIST-GHS-layer1-sparsity}). Layer-2 node sparsity is same in both the $c_{\rm reg}$ values (Fig.~\ref{fig:MLP-MNIST-GHS-layer2-sparsity}). In rest of the experiments involving SS-GHS in main paper, we choose $c_{\rm reg}=1$.
\vspace{0.1in}

%----------------------------------------------------------------------
\section{Definitions and General Lemmas}
\label{AppendixA}
%----------------------------------------------------------------------
\subsection{Definitions}
\begin{definition}\textbf{Hellinger neighborhood:} \label{def:hellinger-neighbor}
We define the $\varepsilon$-Hellinger neighborhood of the true density $P_0$ with 
$\mathcal{H}_{\varepsilon}=\{\btheta: d_{\rm H}(P_0,P_{\btheta})<\varepsilon \}$
where the Hellinger distance, $d_{\rm H}(P_0,P_{\btheta})$, between $P_0$ and the model density function $P_{\btheta}$ is given by 
\begin{equation*}
2d_{\rm H}^2(P_0,P_{\btheta})=\int_{\boldx \in [0,1]^p} \int \left(\sqrt{f_{\btheta}(y|\boldx)}-\sqrt{f_0(y|\boldx)}\right)^2 dy \hspace{0.5mm} d\boldx 
\end{equation*}
\end{definition}

\begin{definition}\textbf{Kullback-Leibler neighborhood:} \label{def:KL-neighbor}
The $\varepsilon$-KL neighborhood of $P_0$ is defined as 
$\mathcal{N}_\varepsilon =\{\btheta: d_{\rm KL}(P_0,P_{\btheta})<\varepsilon \}$
where the KL distance, $d_{\rm KL}(P_0,P_{\btheta})$, between $P_0$ and $P_{\btheta}$ is
\begin{equation*}
d_{\rm KL}(P_0,P_{\btheta})=\int_{\boldx \in [0,1]^p} \int \log \frac{f_0(y|\boldx)}{f_{\btheta}(y|\boldx)}f_0(y|\boldx) \hspace{0.3mm} dy \hspace{0.3mm} d\boldx
\end{equation*}
\end{definition}

\subsection{General Lemmas}
\begin{lemma}
\label{lem:covering-1}
For any 1-Lipschitz continuous (with respect to absolute value) activation function $\psi$, $\abs{\psi(x)}\leq \abs{x}, \hspace{1mm} \forall x \in \mathbb{R}$, %\forall x\geq0$, 
\begin{flalign}
N(\delta,\calF(&L,\bm{k},\bm{s},\bm{B}),||.||_\infty) \leq  \nonumber && \\
& \sum_{s^*_L \leq s_L}\! \cdots \!\sum_{s^*_0 \leq s_0} \left[ \prod_{l=0}^L \left(\frac{B_l k_{l+1}}{\delta B_l/(2(L+1)( \prod_{j=0}^L B_j))} \right)^{s_l } \right] \nonumber &&
\end{flalign}
where $N$ denotes the covering number.
\end{lemma}

\noindent \textbf{Proof} \space Refer to Lemma A.6 in \cite{Jantre-et-al-2023} \hfill $\blacksquare$

%----------------------------------------------------------------------
\section{Requisite Lemmas for Theorems 1 and 3}
\label{AppendixB}
%----------------------------------------------------------------------
\begin{lemma}[Existence of Test Functions]
\label{lem:test}
Let $\epsilon_n \to 0$ and $n \epsilon_n^2 \to \infty$.  There exists a testing function $\phi \in [0,1]$ and constants $C_1,C_2>0$,
\begin{align*}
\E_{P_0}(\phi) &\leq \exp \{-C_1n\epsilon_n^2 \} \\  
\sup_{\substack{\btheta \in \mathcal{H}_{\epsilon_n}^c \\ \eta_{\btheta} \in \mathcal{F}(L,\bm{k},\bm{s}^\circ,\bm{B}^\circ)}}
\E_{P_{\btheta}}(1-\phi)  &\leq \exp \{-C_2nd^2_{\rm H}(P_0,P_{\btheta}) \}
\end{align*}
where $\mathcal{H}_{\epsilon_n}=\{\btheta:d_{\rm H}(P_0,P_{\btheta})\leq \epsilon_n\}$ is the Hellinger neighborhood of radius $\epsilon_n$.
\end{lemma}
\noindent \textbf{Proof} \space Let $H=N(\epsilon_n/19,\calF(L,\bm{k},\bm{s}^\circ,\bm{B}^\circ),d_{\rm H}(.,.))$ denote the covering number of the $\calF(L,\bm{k},\bm{s}^\circ,\bm{B}^\circ)$. 

Using Lemma \ref{lem:covering-1} with $\bm{s}=\bolds^\circ$ and $\bm{B}=\boldB^\circ$, we get
\begin{align}
&\log H  = \log N(\epsilon_n/19,\mathcal{F}(L,\bm{k},\bm{s}^\circ,\bm{B}^\circ),d_{\rm H}(.,.)) \nonumber \\
&\leq \log N(\sqrt{8}\sigma^2_e \epsilon_n/19,\calF(L,\bm{k},\bm{s}^\circ,\bm{B}^\circ),||.||_\infty) \nonumber\\
& \leq \log \left[ \prod_{l=0}^L \left(\frac{38}{\sqrt{8}\sigma^2_e \epsilon_n}(L+1)  \prod_{j=0}^L B_j^\circ k_{l+1}\right)^{(s_{l}^\circ+1)} \right]\nonumber \\
& \leq C \left[\sum_{l=0}^L (s_{l}^\circ+1)\left(\log \left(\frac{1}{\epsilon_n} (L+1) \right) +  \sum_{j=0}^L \log B_j^\circ + \log k_{l+1}\right) \right]\nonumber \\
&\leq  C \sum_{l=0}^L (s_{l}^\circ+1) (\log n+\log (L+1)+\sum_{j=0}^{L} \log B_j^\circ+\log k_{l+1}) \nonumber\\
&\leq C\sum_{l=0}^L (s_l^\circ+1)(\log n +\log (L+1) \nonumber \\
& \qquad + \sum_{l=0}^L \log {k_{l+1}}+\sum_{l=0}^L \log (k_l+1) + A)  \leq Cn\epsilon_n^2  \nonumber
\end{align}
{\bf Group Lasso Case:} $A = -\log \min(1,\varsigma^2)$. Next, let $B_l^\circ=2(n\epsilon_n^2+\log (k_l+1)+\log k_{l+1}+\log(L+1)+\log 4)(k_l+1)/\min(1,\varsigma^2)$ and $s_l^\circ+1=n\epsilon_n^2/\sum_{l=0}^Lu_l$ where $u_l=\log n+\log (L+1)+\sum_{l=0}^L \log k_l+\sum_{l=0}^L \log (k_l+1)$.
\vspace{0.1in}

\noindent {\bf Group Horseshoe Case:} $A = \log c^2_{\rm reg}$. Next, let  $(B_l^\circ)^2=2(n\epsilon_n^2+\log (k_l+1)+\log k_{l+1}+\log(L+1)+\log 2)(k_l+1)^2c^2_{\rm reg}$ and $s_l^\circ+1=n\epsilon_n^2/\sum_{l=0}^Lu_l$ where $u_l=\log n+\log (L+1)+\sum_{l=0}^L \log k_l+\sum_{l=0}^L \log (k_l+1)+\log c^2_{\rm reg}$.
\vspace{0.03in}

\noindent Remaining steps are same as Proof of Lemma 4.1 in \cite{Jantre-et-al-2023}. 
\hfill $\blacksquare$

\begin{lemma}[Prior mass condition] 
\label{lem:prior} 
Let $\epsilon_n \to 0$, $n \epsilon_n^2 \to \infty$ and  $n\epsilon_n^2/\sum_{l=0}^L u_{l} \to \infty$, then for prior $\widetilde{\Pi}$  and some constant $C_3>0$, $ \widetilde{\Pi}(\mathcal{F}(L,\bm{k},\bm{s}^\circ,\bm{B}^\circ)^c)\leq \exp (-C_3n\epsilon_n^2/\sum_{l=0}^L u_l)$.
\end{lemma}
\noindent \textbf{Proof} \space  We provide proof for SS-GL in supplement Appendix A and SS-GHS in supplement Appendix B. \hfill $\blacksquare$
 
Lemma~\ref{lem:kl} has two conditions. Condition 1 requires that shrinking KL neighborhoods of the true density function $P_0$ get a sufficiently large probability to guarantee convergence of the true posterior in (5). Condition 2. helps control the KL distance between true posterior and variational posterior to guarantee the convergence of the variational posterior in (6). 

\begin{lemma}[Kullback-Leibler conditions]
\label{lem:kl}
Suppose $\sum_{l=0}^L r_l+\xi \to 0$ and $n (\sum_{l=0}^L r_l+\xi) \to \infty$. Assume following two conditions hold for the prior $\widetilde{\Pi}$ and  some $q \in \mathcal{Q}$:\\
(1) $\widetilde{\Pi}\left(\mathcal{N}_{\sum_{l=0}^L r_l+\xi}\right)\geq \exp(-C_4 n(\sum_{l=0}^L r_l+\xi))$,\\
(2) $d_{\rm KL}(q,\pi)+n\sum\limits_{\boldz} \int d_{\rm KL}(P_0,P_{\btheta})q(\btheta,\boldz)d\btheta\leq C_5 n(\sum_{l=0}^L r_l+\xi)$ where $\mathcal{N}_{\sum_{l=0}^L r_l+\xi} $ be the KL neighborhood of radius $\sum_{l=0}^L r_l +\xi$.
\end{lemma}
\noindent \textbf{Proof} \space 
% The proof of this lemma differs in SS-GL and SS-GHS. 
We provide proof for SS-GL in supplement Appendix A and SS-GHS in supplement Appendix B. \hfill $\blacksquare$

\begin{lemma}
\label{lem:covering}
If Lemma \ref{lem:test} and \ref{lem:prior} hold, then with dominating probability, $$\log \int_{\mathcal{H}_{\epsilon_n}^c}(P_{\btheta}^n/P_0^n)\pi(\btheta) d\btheta \leq -C n{\epsilon_n^2}/\sum u_l.$$
\end{lemma}
\noindent \textbf{Proof} \space Refer to Lemma A.8 in \cite{Jantre-et-al-2023}. \hfill $\blacksquare$

\begin{lemma}
\label{lem:kl-denominator}
If Lemma \ref{lem:kl} part 1 holds, $\forall M_n \to \infty$, with dominating probability, $ \log  \int (P_0^n/P_{\btheta}^n)\widetilde{\pi}(\btheta)d\btheta\leq n M_n(\sum r_l+\xi)$.
\end{lemma}
\noindent \textbf{Proof} \space Refer to Lemma A.9 in \cite{Jantre-et-al-2023}. \hfill $\blacksquare$

\begin{lemma}
\label{lem:q-determination}
If Lemma \ref{lem:kl} part 2 holds, for any $M_n \to \infty$, with dominating probability,\\
$d_{\rm KL}(q,\pi)+\sum_{\boldz}\int \log (P_0^n/P_{\btheta}^n) q(\btheta,\boldz)d\btheta \leq n M_n(\sum r_l+\xi)$.
\end{lemma}
\noindent \textbf{Proof} \space Refer to Lemma A.10 in \cite{Jantre-et-al-2023}. \hfill $\blacksquare$

%----------------------------------------------------------------------
\section{Spike-and-Slab Group Lasso Proofs}
\label{AppendixC}
%----------------------------------------------------------------------
\subsection{ELBO derivation}
\label{appendix:elbo-ssgl}
\begin{align*}
\mathcal{L} & = -\E_{q(\btheta,\bupvarpi)}[\log L(\btheta)] + \sum_{l,j} q(\boldz_{lj} = 1)  \int d_{\rm KL}( N(\bmu_{lj},\text{diag}(\bsigma^2_{lj})), N(0,\sigma_0^2 \tau^{2}_{lj} \boldI)) q(\tau^{2}_{lj}) d \tau^{2}_{lj} \\
& \quad + \sum_{l,j} d_{\rm KL}(\text{Ber}(\gamma_{lj}),\text{Ber}(\lambda_l)) + \sum_{l,j} \int d_{\rm KL}( LN(\mu_{lj}^{\{\tau\}},\sigma^{\{\tau\}2}_{lj}),  G(k_l/2+1,\varsigma^2/2)) q(\varsigma^2)d\varsigma^2\\
& \quad +LN(\mu^{\{\varsigma\}},\sigma^{\{\varsigma\}2}_{lj}),  G(a_0,b_0))
\end{align*}

\subsection{Proof of Lemma 11.}
\noindent Let us start with the following assumption:
\begin{align}
\label{e:ass-lem-prior}
{\it Assumption:}  \hspace{3mm} s_{l}^\circ+1 = (n\epsilon_n^2 )/(\sum u_l ),\hspace{2mm} \lambda_l k_{l+1}/s_l^\circ \to 0,\hspace{2mm} \sum u_l\log L=o(n\epsilon_n^2)
\end{align}
\begin{align*}
\widetilde{\Pi}(\mathcal{F}(L,\bm{k},\bolds^\circ,\boldB^\circ)^c) &\leq \widetilde{\Pi}\left(\bigcup_{l=0}^L \{ ||\widetilde{\boldw}_l||_0>s_l^\circ\} \right) + \widetilde{\Pi}\left(\bigcup_{l=0}^L \{ ||\widetilde{\boldw}_l||_\infty > B_l^\circ \} \right) \\
& \leq \sum_{l=0}^L \widetilde{\Pi}(||\widetilde{\boldw}_l||_0>s_l^\circ) + \sum_{l=0}^L \widetilde{\Pi}(||\widetilde{\boldw}_l||_\infty > B_l^\circ) \\
& = \sum_{l=0}^L \int \int \sum_{\boldz}  \Pi(||\widetilde{\boldw}_l||_0 > s_l^\circ |\btau^2,\boldz) \pi(\boldz) \pi(\btau^2)d\btau^2 \\
& \quad + \sum_{l=0}^L \int \int \sum_{\boldz} \Pi(||\widetilde{\boldw}_l||_\infty > B_l^\circ|\btau^2,\boldz)\pi(\boldz) \pi(\btau^2)d\btau^2 \\
& \leq \sum_{l=0}^L \prob(\sum_{i=1}^{k_{l+1}} z_{li}>s_{l}^\circ) + \sum_{l=0}^L \int \int \prob( \sup_{i=1,\cdots,k_{l+1}} ||\overline{\boldw}_{li}||_2 > B_l^\circ|\btau^2)\pi(\btau^2)d\btau^2
\end{align*}
where $\widetilde{\boldw}_l=(||\overline{\boldw}_{l1}||_2,\cdots,|| \overline{\boldw}_{lk_{l+1}}||_2)^T$ and the last inequality holds since for the first part, $\Pi(||\widetilde{\boldw}_l||_0>s_l^\circ|\btau^2,\boldz) \leq 1$, $\Pi(||\widetilde{\boldw}_l||_0>s_l^\circ|\btau^2,\boldz)=1$ iff $\sum z_{li}>\bolds_l^\circ$ and for the second part, $\pi(\boldz)\leq 1$.

\vspace{0.1in}
\noindent \underline{\it Part 1.} Check the proof of Lemma 4.2 in \cite{Jantre-et-al-2023} by which 
$$\sum_{l=0}^L \prob(\sum_{i=1}^{k_{l+1}} z_{li}>s_{l}^\circ )= \sum_{l=0}^L \prob(\sum_{i=1}^{k_{l+1}} z_{li}-k_{l+1}\lambda_l>s_{l}^\circ-k_{l+1}\lambda_l)$$.

\noindent \underline{\it Part 2.} \vspace{-2mm}
\begin{align*}
&\sum_{l=0}^L \int \prob( \sup_{i=1,\cdots,k_{l+1}} ||\boldw_{li}||_2 > B_l^\circ|\btau^2)\pi(\btau^2)d\btau^2 \\
&\leq \sum_{l=0}^L \sum_{i=1}^{k_{l+1}} \int \int \prob\left( ||\boldw_{li}||_2 > B_l^\circ | \tau_{li}^2 \right) \pi(\tau_{li}^2|\varsigma^2)\pi(\varsigma^2) d\tau_{li}^2 d\varsigma^2\\ 
&\leq \sum_{l=0}^L \sum_{i=1}^{k_{l+1}}   \sum_{j=1}^{k_l+1}\int \prob\Big(|w_{lij}|>\frac{B_l^\circ}{k_{l}+1}\Big|\tau_{li}^2 \Big) \pi(\tau_{li}^2|\varsigma^2)\pi(\varsigma^2) d\tau_{li}^2 d\varsigma^2\\
&\leq 2\sum_{l=0}^L \sum_{i=1}^{k_{l+1}} \sum_{j=1}^{k_l+1} \int \int \exp\Big(-\frac{{B_l^\circ}^2}{2(k_l+1)^2\tau_{li}^2}\Big) \pi(\tau_{li}^2|\varsigma^2) \pi(\varsigma^2) d\tau_{li}^2  d\varsigma^2 \hspace{5mm} \text{By  concentration inequality} \\
&\leq 2 \sum_{l=0}^L \sum_{i=1}^{k_{l+1}} \sum_{j=1}^{k_l+1} \int \Big[ \exp\Big(-\frac{\varsigma^2{B_l^\circ}}{2(k_l+1)}\Big) \prob\Big(\tau_{li}^2 \leq \frac{B_l^\circ} {\varsigma^2(k_l+1)}\Big) +  \prob\Big(\tau_{li}^2 \geq \frac{B_l^\circ}{\varsigma^2(k_l+1)}\Big) \Big] \pi(\varsigma^2)d\varsigma^2 \\
&\leq  2\sum_{l=0}^L \sum_{i=1}^{k_{l+1}} \sum_{j=1}^{k_l+1}  \int \Big[\exp\Big(-\frac{\varsigma^2{B_l^\circ}}{2(k_l+1)}\Big)+\exp\Big(-\frac{B_l^\circ}{k_l+1}\Big)\Big] \pi(\varsigma^2)d\varsigma^2  \\
&\lesssim  4\sum_{l=0}^L \sum_{i=1}^{k_{l+1}} \sum_{j=1}^{k_l+1} \exp\Big(\frac{-B_l^\circ}{2(k_l+1)}\Big) 
\leq \sum_{l=0}^L \sum_{i=1}^{k_{l+1}} \sum_{j=1}^{k_l+1} \frac{1}{(L+1) k_{l+1}(k_l+1)}\exp(-n\epsilon_n^2) 
\end{align*}
where the third  inequality holds since $|w_{lij}|$ given $\tau_{li}^2$ is bound above by a $|N(0,\tau_{li}^2\sigma_0^2)|$ random variable and we take $\sigma_0^2=1$ w.l.o.g and the sixth inequality holds because $C=\min(1,a_0/b_0)=1$ w.l.o.g. and
\begin{align*}
  \int\exp\Big(-\frac{\varsigma^2{B_l^\circ}}{2(k_l+1)}\Big)  \pi(\varsigma^2)d\varsigma^2  &=\frac{b_0^{a_0}}{\Gamma(a_0)}\int (\varsigma^2)^{a_0-1} \exp\Big(-\varsigma^2\Big(\frac{{B_l^\circ}}{2(k_l+1)}+b_0\Big)\Big) d\varsigma^2 \\
  &=b_0^{a_0}/\Big(\frac{B_l^\circ}{2(k_l+1)}+b_0\Big)^{a_0} = \Big(1+\frac{B_l^\circ}{2(k_l+1)b_0}\Big)^{-a_0} \sim \exp\Big(-\frac{B_l^\circ a_0}{2(k_l+1)b_0}\Big)
\end{align*}
provided $\frac{B_l^\circ}{k_l+1} \to \infty $ which is true when  $B_l^\circ=2(n\epsilon_n^2+\log (k_l+1)+\log k_{l+1}+\log(L+1)+\log 4)(k_l+1)$.
\hfill $\blacksquare$

\subsection{Proof of Lemma 12 Condition 1.}
\noindent Let us start with the following assumption:
\begin{align}
\label{e:ass-lem-kl-ss-gl}
{\it Assumption:} & -\log \lambda_l =O((k_l+1) \vartheta_l), \hspace{3mm} -\log(1-\lambda_l) = O(s_l(k_l+1)\vartheta_l/k_{l+1} )
\end{align}
Let $\sigma^2_e = 1$ w.l.o.g. Then $P_0(y,\boldx)  =\exp (- (y-\eta_0(\boldx))^2 )/\sqrt{2\pi}$ and 
$P_{\btheta}(y,\boldx)  = \exp (-(y-\eta_{\btheta}(\boldx))^2))/\sqrt{2\pi }$ which implies
\begin{align}
\label{e:kl-p0-pt}
  2d_{\text{KL}}(P_0,P_{\btheta}) 
= \int_{\boldx\in[0,1]^p} (\eta_0(\boldx)-\eta_{\btheta}(\boldx))^2 d\boldx=||\eta_0-\eta_{\btheta}||_2^2
\end{align}
Next, let $\eta_{\btheta^*}(\boldx)$  minimize ${\eta_{\btheta} \in \calF(L,\boldk,\bolds,\boldB)} \left\|\eta_{\btheta}-\eta_{0}\right\|_{\infty}^{2}$ 
\begin{equation} \label{first_diff}
    ||\eta_{\btheta^*}-\eta_0||_1 \leq ||\eta_{\btheta^*}-\eta_0||_\infty = \sqrt{\xi}
\end{equation}
Consider $L_2$ norm of rows of $\overline{\boldD}_l = \overline{\boldW}_l - \overline{\boldW}^*_l$, i.e. $\overline{\boldD}_l=(\overline{\bm{d}}_{l1}^\top,..,\overline{\bm{d}}_{lk_{l+1}}^\top)^{\top} $ and define $ \overline{\bdelta}_l=(||\overline{\bm{d}}_{l1}||_2 ,..,||\overline{\bm{d}}_{lk_{l+1}}||_2 )$. $\mathcal{S}_l^c$ is the set where $||\overline{\boldw}_{li}^*||_2=0$, $l=0,\cdots, L$. Define $\calM_{\sqrt{\sum r_l} }$ as: $$ \calM_{\sqrt{\sum r_l}} =\left\{\btheta: ||\overline{\bm{d}}_{li}||_2 \leq \frac{\sqrt{\sum r_l}B_l}{(L+1)(\prod_{j=0}^L B_j)}, i\in \mathcal{S}_l, ||\overline{\bm{d}}_{li}||_2=0, i \in \mathcal{S}_l^c, l = 0,\cdots,L  \right\}$$
 For every $\btheta \in \calM_{\sqrt{\sum r_l}}$ by relation (16) in \cite{Jantre-et-al-2023},
 \begin{equation} \label{second_diff}
    || \eta_{\btheta} - \eta_{\btheta^*} ||_1 \leq \sqrt{\sum r_l} 
\end{equation}
Let $\delta_n = (\sqrt{\sum r_l}B_l)/((L+1)(\prod_{j=0}^L B_j))$, $A = \{\overline{\boldw}_{li} : ||\overline{\boldw}_{li} - \overline{\boldw}_{li}^*||_2 \leq \delta_n \}$. By \eqref{first_diff} and \eqref{second_diff}, for $\btheta \in \calM_{\sqrt{\sum r_l}}, ||\eta_{\btheta}-\eta_0||_1 \leq \sqrt{\sum r_l}+ \sqrt{\xi}.$ 
$d_{\text{KL}}(P_0,P_{\btheta}) \leq (\sqrt{\sum r_l}+ \sqrt{\xi})^2/2\leq \sum r_l+ \xi$. Since $\btheta \in \mathcal{N}_{\sum r_l+\xi}$ for every $\btheta \in \calM_{\sqrt{\sum r_l}}$, $$\int_{\btheta \in \mathcal{N}_{\sum r_l + \xi}} \widetilde{\pi}(\btheta)d\btheta \geq \int_{\btheta \in \calM_{\sqrt{\sum r_l}}} \widetilde{\pi}(\btheta) d\btheta.$$
\begin{align*}
& \vspace{-2mm} \widetilde{\Pi}\left(\calM_{\sqrt{\sum r_l}}\right) \\
&= \int  \int \sum_{\boldz}\Pi\left(\calM_{\sqrt{\sum r_l}}\Big|\btau^2, \boldz\right) \pi(\boldz) \pi(\btau^2|\varsigma^2) \pi(\varsigma^2)  d\btau^2 d\varsigma^2\\ 
&\geq \int \int \sum_{\{\boldz: z_{li}=1, i \in \mathcal{S}_l, z_{li}=0, i \in \mathcal{S}_l^c, l=0,\cdots, L\}} \Pi\left(\calM_{\sqrt{\sum r_l}}\Big|\btau^2,\boldz\right)\pi(\boldz) \pi(\btau^2|\varsigma^2) d\btau^2 d\varsigma^2\\ 
&= \prod_{l=0}^L  (1-\lambda_l)^{k_{l+1}-s_l} \lambda_l^{s_l}  \prod_{i \in \mathcal{S}_l}\int \int \E (\indicator_{\{\overline{\boldw}_{li} \in A \}}|\tau_{li}^2,z_{li}=1) \pi(\tau_{li}^2|\varsigma^2) \pi(\varsigma^2) d\tau_{li}^2 d\varsigma^2\\
&\geq  \prod_{l=0}^L  (1-\lambda_l)^{k_{l+1}-s_l} \lambda_l^{s_l} \prod_{i \in \mathcal{S}_l} \int \E (\indicator_{\{\overline{\boldw}_{li} \in A \}}|\tau_{li}^2=(k_l+1)/\varsigma^2,z_{li}=1) \prob(\tau_{li}^2 \leq (k_l+1)/\varsigma^2) \pi(\varsigma^2) d\varsigma^2 \\
&= \prod_{l=0}^L (1-\lambda_l)^{k_{l+1}-s_l} \lambda_l^{s_l} \prod_{i \in \mathcal{S}_l} \int  \Bigg[\int_{\overline{\boldw}_{li} \in A} \left(\frac{\varsigma^2}{2\pi(k_l+1) }\right)^{\frac{k_l+1}{2}}\prod_{j=1}^{k_l+1} \exp \left(- \frac{\varsigma^2\overline{w}_{lij}^2}{2(k_l+1)} \right) d \overline{w}_{lij} \Bigg] \pi(\varsigma^2)d\varsigma^2\\
& \geq \prod_{l=0}^L (1-\lambda_l)^{k_{l+1}-s_l} \lambda_l^{s_l} \prod_{i \in \mathcal{S}_l}  \int  \Bigg[ \left(\frac{\varsigma^2}{2\pi(k_l+1)}\right)^{\frac{k_l+1}{2}}\prod_{j=1}^{k_l+1} \int_{\overline{w}^*_{lij}-\frac{\delta_n}{\sqrt{k_l+1}}}^{\overline{w}^*_{lij}+\frac{\delta_n}{\sqrt{k_l+1}}} \exp \left(- \frac{\varsigma^2\overline{w}_{lij}^2}{2(k_l+1)} \right) d \overline{w}_{lij}\Bigg] \pi(\varsigma^2)d\varsigma^2  \\
& \geq \prod_{l=0}^L (1-\lambda_l)^{k_{l+1}-s_l} \lambda_l^{s_l}  \prod_{i \in \mathcal{S}_l}  \int  \Bigg[ \left(\frac{\varsigma^2}{2\pi(k_l+1)}\right)^{\frac{k_l+1}{2}} \prod_{j=1}^{k_l+1} \frac{2\delta_n}{\sqrt{k_l+1}}\exp \left(- \frac{\widehat{w}_{lij}^2 \varsigma^2}{2(k_l+1)} \right) \Bigg] \pi(\varsigma^2)d\varsigma^2
\end{align*}
where second equality holds by  $\E(\indicator_{\{\overline{\boldw}_{li} \in A \}}| \tau_{li}^2=(k_l+1)/ \varsigma^2, z_{li}=0)=1 $, $||\overline{\boldw}_{li}^*||_2=0$, for $i \in \mathcal{S}_l^c$. Last equality is by mean value theorem and $\widehat{w}_{lij} \in [\overline{w}^*_{lij}-\delta_n /(k_l+1),\overline{w}^*_{lij}+\delta_n/(k_l+1)]$.
Simplifying further we get
\begin{align*}
\widetilde{\Pi}\left(\calM_{\sqrt{\sum r_l}}\right)    & \geq \prod_{l=0}^L (1-\lambda_l)^{k_{l+1}-s_l} \lambda_l^{s_l}  \prod_{i \in \mathcal{S}_l} \Big(\frac{2\delta_n}{\sqrt{2\pi}k_l+1}\Big)^{k_l+1}  \int  \Bigg[ (\varsigma^2)^{\frac{k_l+1}{2}}  \exp \left(- \frac{\sum_{j=1}^{k_l+1}\widehat{w}_{lij}^2 \varsigma^2}{2(k_l+1)} \right) \Bigg] \pi(\varsigma^2)d\varsigma^2\\
&=\prod_{l=0}^L (1-\lambda_l)^{k_{l+1}-s_l} \lambda_l^{s_l}  \prod_{i \in \mathcal{S}_l} \Big(\frac{2\delta_n}{\sqrt{2\pi}k_l+1}\Big)^{k_l+1}  \int  \Bigg[ (\zeta^2)^{-\frac{k_l+1}{2}}\exp \left(-  \frac{\sum_{j=1}^{k_l+1} 
 \widehat{w}_{lij}^2 }{2\zeta^2(k_l+1)} \right) \Bigg] \pi(\zeta^2)d\zeta^2\\
&\gtrsim \prod_{l=0}^L (1-\lambda_l)^{k_{l+1}-s_l} \lambda_l^{s_l}  \prod_{i \in \mathcal{S}_l} \Big(\frac{2\delta_n}{\sqrt{2\pi}k_l+1}\Big)^{k_l+1}  \int_{t_0''}  \Bigg[ (\zeta^2)^{-\frac{k_l+1}{2}-a''-1}  \exp \left(- \frac{\sum_{j=1}^{k_l+1}\widehat{w}_{lij}^2 }{2\zeta^2(k_l+1)} \right) \Bigg] d\zeta^2\\
&\gtrsim \prod_{l=0}^L (1-\lambda_l)^{k_{l+1}-s_l} \lambda_l^{s_l}  \prod_{i \in \mathcal{S}_l} \Big(\frac{2\delta_n}{\sqrt{2\pi}k_l+1}\Big)^{k_l+1}  \exp \left(- \frac{\sum_{j=1}^{k_l+1}\widehat{w}_{lij}^2 }{2t_0''(k_l+1)} \right)\int_{t_0''}  (\zeta^2)^{-\frac{k_l+1}{2}-a''-1}   d\zeta^2\\
& =\prod_{l=0}^L (1-\lambda_l)^{k_{l+1}-s_l} \lambda_l^{s_l}  \prod_{i \in \mathcal{S}_l} \Big(\frac{2\delta_n}{\sqrt{2\pi}k_l+1}\Big)^{k_l+1} \frac{t_0''^{-\frac{k_l+1}{2}-a''}}{\frac{k_l+1}{2}+a''}   \exp \left(- \frac{\sum_{j=1}^{k_l+1}\widehat{w}_{lij}^2 }{2t_0''(k_l+1)} \right)
\end{align*}
where $\zeta=1/\varsigma$ and follows Inverse Gamma distribution with parameters. Simplifying further
\begin{align}
\nonumber \widetilde{\Pi}\left(\calM_{\sqrt{\sum r_l}}\right) &\gtrsim  \exp\Bigg[ -\sum_{l=0}^L\Bigg\{ s_l \log \left( \frac{1}{\lambda_l}\right)+(k_{l+1}-s_l)\log\left(\frac{1}{1-\lambda_l}\right) \\
& \nonumber + \sum_{i\in \mathcal{S}_l} \Bigg( \frac{k_l+1}{2} \log (2\pi(k_l+1)) - (k_l+1) \log \frac{2\delta_n }{\sqrt{k_l+1}} +  \sum_{j=1}^{k_l+1} \frac{\widehat{w}_{lij}^2}{2t_0''
(k_l+1)} \\
& \nonumber \qquad \qquad \qquad + \frac{(k_l+1) }{2}\log t_0''+\log\Big(\frac{(k_l+1) }{2}+a''\Big) \Bigg) \Bigg\}  \Bigg]\\
& \nonumber \gtrsim \exp\Bigg[ -\sum_{l=0}^L\Bigg\{ s_l \log \left( \frac{1}{\lambda_l}\right)+(k_{l+1}-s_l)\log\left(\frac{1}{1-\lambda_l}\right)+ \frac{s_l(k_l+1)}{2}(\log t_0''+\log (2\pi))\\
& \nonumber \quad + s_l(k_l+1) \left(\frac{1}{2}\log(k_l+1) + \frac{1}{2}\log (k_l+1)-\log (2\delta_n)\right) \\
& \qquad \qquad \qquad + \sum_{i\in \mathcal{S}_l}  \sum_{j=1}^{k_l+1} \frac{\widehat{w}_{lij}^2 }{2(k_l+1)t_0''} +\frac{\log (k_l+1)}{(k_l+1)}\Bigg\}  \Bigg] \label{ss_gl_KL_lemma_main_expression}
\end{align}
Using $\delta_n \to 0$, we get
\begin{align}
\label{ss_gl_KL_lemma_weight_bound}
\nonumber  & \frac{1}{2t_0''} \sum_{l=0}^L \sum_{i\in \mathcal{S}_l} \sum_{j=1}^{k_l+1} \frac{\widehat{w}_{lij}^2}{k_l+1}   \leq \frac{1}{2t_0''}\sum_{l=0}^L \sum_{i\in \mathcal{S}_l} \sum_{j=1}^{k_l+1} \frac{(\max ((\overline{w}^*_{lij}-\delta_n/\sqrt{k_l+1}), (\overline{w}^*_{lij}+\delta_n/\sqrt{k_l+1})))^2}{k_l+1} \\
 & \leq \frac{1}{t_0''
 }\sum_{l=0}^L \sum_{i\in \mathcal{S}_l} \sum_{j=1}^{k_l+1} \frac{\overline{w}_{lij}^{*2} + \delta_n^2/(k_{l}+1)}{k_l+1}
\leq \sum_{l=0}^L s_l(k_l+1) \left(\frac{B_l^2}{(k_l+1)} + \frac{1}{(k_l+1)} \right)\frac{1}{t_0''(k_l+1)} 
\end{align}
and  
\begin{flalign}
\nonumber&\sum_{l=0}^L \Bigg( s_l \log \Big( \frac{1}{\lambda_l}\Big)+(k_{l+1}-s_l)\log\Big(\frac{1}{1-\lambda_l}\Big) + \frac{s_l(k_l+1)}{2} \Big(\log (k_l+1)^2-2\log (2  \delta_n\sqrt{2\pi e})+\log t_0'' \Big)\Bigg)\\
\nonumber & \sim \sum_{l=0}^L \Bigg( s_l \log \Big( \frac{1}{\lambda_l}\Big)+(k_{l+1}-s_l)\log\Big(\frac{1}{1-\lambda_l}\Big)\Bigg)\\
\nonumber &+\sum_{l=0}^L \frac{s_l(k_l+1)}{2} \Big(2\log (k_l+1)+\log t_0''+2\log (L+1)  + 2\sum\log B_m-\log B_l^2-\log \sum r_l \Big)\nonumber\\
& \leq \sum_{l=0}^L C n r_l+\sum_{l=0}^L\frac{s_l(k_l+1)}{2}\Big(-\log \frac{B_l^2 }{t_0''(k_l+1)^2}+\log (L+1)+2\sum \log B_m+  \log n \Big)\leq \sum_{l=0}^L Cnr_l
\label{ss_gl_KL_lemma_rest_bound}
\end{flalign}
where the first inequality follows from \eqref{e:ass-lem-kl-ss-gl} and expanding $\delta_n$. The last inequality follows since $n\sum r_l \to \infty$. Combining \eqref{ss_gl_KL_lemma_weight_bound} and \eqref{ss_gl_KL_lemma_rest_bound} gives the bound for \eqref{ss_gl_KL_lemma_main_expression}. \hfill $\blacksquare$

\subsection{Proof of Lemma 12 Condition 2.}
\label{lemma:KL_likelihood_bound_lemma}

\noindent Let us start with the following assumption:
\begin{align*}
{\it Assumption:} -\log \lambda_l=O\{(k_l+1) \vartheta_l\},\: -\log(1-\lambda_l) = O\{ (s_l/k_{l+1})(k_l+1) \vartheta_l\}
\end{align*}

\noindent Suppose there exists $q^* \in \mathcal{Q}$ such that
\begin{eqnarray}
\label{KL_ineq} d_{\rm KL}(q^*,\pi) \leq C_1 n \sum r_l \hspace{5mm}
\sum_{\boldz}\int_{\bTheta} \left\|\eta_{\btheta}-\eta_{\btheta^*}\right\|_{2}^{2} q^*(\btheta,\boldz)d\btheta \leq \sum r_l. \label{inf_norm_ineq}
\end{eqnarray}
Recall $\btheta^* = \arg \min_{\theta:\eta_{\btheta} \in \mathcal{F}(L,p,\bolds,\boldB)} \left\|\eta_{\btheta}-\eta_{0}\right\|_{\infty}^{2}$. By 
 relation \eqref{e:kl-p0-pt},
\begin{align*}
\sum_{\boldz}\int nd_{\rm KL}(P_0,P_{\btheta})q^*(\btheta,\boldz)d\btheta&
\leq \frac{n}{2}(\sum_{\boldz}\int ||\eta_{\btheta^*}-\eta_{\btheta}||_2^2q^*(\btheta,\boldz)d\btheta+\frac{n}{2}||\eta_{\btheta^*}-\eta_0||_\infty^2\leq C n (\sum r_l+\xi)
\end{align*}
where the above relation is due to \eqref{inf_norm_ineq}. Next, we construct $q^*$ as
$$\begin{aligned}
& \overline{w}_{lij}|z^*_{li} \sim z^*_{li} \calN (\overline{w}_{lij}^*,\sigma_l^2) + (1-z^*_{li}) \delta_0, \qquad z^*_{li} \sim \text{Bern} (\gamma_{li}^*) 
\qquad \gamma_{li}^* =  \indicator(||\boldw_{li}^*||_2 \neq 0)
\end{aligned}$$
where $\sigma_l^2 = s^*_{l}/(8n(L+1)(4^{L-l}  (k_{l}+1)\log( s^*_{l} 2^{k_{l}+1}) \prod_{m=0,m\neq l}^{L} B_m^2)).$  For this choice of $q^*$, we have
$\int ||\eta_{\btheta} - \eta_{\btheta^*}||^2_2 q^*(\btheta,\boldz)d\btheta  \leq \sum r_l$. The proof of this follows as a consequence of Proof of Lemma 4.3 part 2. in \cite{Jantre-et-al-2023}. Next, we upper bound $d_{\rm KL}(q^*,\pi)$ by
\begin{align*}
&\log\frac{1}{\pi(\boldz^*)} +  \indicator (\boldz=\boldz^*) \int d_{\rm KL}\Bigg[\Bigg(\prod_{l=0}^{L-1} \prod_{i=1}^{k_{l+1}} \prod_{j=1}^{k_{l}+1} \{ z_{li}\calN(\overline{w}_{lij}^*,\sigma_l^2) +(1-z_{li})\delta_0 \}  \prod_{j=1}^{k_{L}+1} \calN (\overline{w}_{Lj}^*,\sigma_L^2) \Bigg),\\
&\hspace{3cm}\Bigg( \prod_{l=0}^{L-1} \prod_{i=1}^{k_{l+1}} \prod_{j=1}^{k_{l}+1} \{ z_{li}\calN(0,\sigma^2_0\tau_{li}^{2}) +(1-z_{li})\delta_0 \} \prod_{j=1}^{k_{L}+1} \calN (0,\sigma^2_0\tau_{L}^{2}) \Bigg) \Bigg]  \prod_{l,i} LN\left(\mu_{li}^{\{\tau\}},\sigma^{\{\tau\}2}_{li}\right) d \tau^{2}_{li}\\
& + \int  \Bigg[\sum_{l,i} d_{\rm KL}\left( LN\left(\mu_{li}^{\{\tau\}},\sigma^{\{\tau\}2}_{li}\right),  \Gamma\left(\frac{k_l+2}{2},\frac{\varsigma^2}{2}\right)\right)\Bigg] LN(\mu^{\{\varsigma\}},\sigma^{\{\varsigma\}2}) d\varsigma^2 \\
& + d_{\rm KL} \left(LN\left(\mu_{li}^{\{\varsigma\}},\sigma^{\{\varsigma\}2}_{li}\right),  \Gamma(a_0,b_0)\right)\\
& = \sum_{l=0}^{L-1} \Big[-s^*_{l} \log \lambda_l- (k_{l+1}-s^*_{l}) \log (1-\lambda_l) \Big] \\
& + \sum_{l=0}^{L-1} \sum_{i=1}^{k_{l+1}} z_{li} \sum_{j=1}^{k_{l}+1} \Bigg\{\frac{1}{2}\log \frac{\sigma^2_0}{\sigma_l^2}+\frac{1}{2}\mu_{li}^{\{\tau\}}+  \frac{\sigma_l^2+{\overline{w}_{lij}^*}^2}{2\sigma^2_0}\exp\left( -\mu_{li}^{\{\tau\}} + \frac{\sigma^{\{\tau\}2}_{li}}{2}\right) - \frac{1}{2} \Bigg\} \\
& + \sum_{j=1}^{k_{L}+1} \Bigg\{\frac{1}{2}\log \frac{\sigma^2_0}{\sigma_L^2} + \frac{1}{2}\mu_{L}^{\{\tau\}} + \frac{\sigma_L^2+{\overline{w}_{Lj}^*}^2}{2\sigma^2_0} \exp \left( -\mu_{L}^{\{\tau\}} + \frac{\sigma^{\{\tau\}2}_{L}}{2} \right) - \frac{1}{2} \Bigg\} \\
& + \sum_{l=0}^{L} \sum_{i=1}^{k_{l+1}} \left[\frac{k_l+2}{2}(\log2-\mu_{li}^{\{\tau\}}-\mu^{\{\varsigma\}}) + \log \Gamma\left(\frac{k_l+2}{2}\right)   + \frac{1}{2} \exp\left(\mu_{li}^{\{\tau\}} + \frac{\sigma^{\{\tau\}2}_{li}}{2}+\mu^{\{\varsigma\}} + \frac{\sigma^{\{\varsigma\}2}}{2}\right) \right. \\
& \hspace{6cm} \left.- \log \sigma^{\{\tau\}}_{li} - \log \sqrt{2\pi} - \frac{1}{2} \right]\\
& + \log \Gamma(a_0)-a_0(b_0+\mu^{\{\varsigma\}})+b_0 \exp\left(\mu^{\{\varsigma\}}+
\frac{\sigma^{\{\varsigma\}}}{2}\right)-\log \sigma^{\{\varsigma\}} -\log \sqrt{2\pi}-\frac{1}{2}\\
&\leq \sum_{l=0}^{L-1} C n r_l +  \sum_{l=0}^{L-1} \sum_{i=1}^{s^*_{l}} \frac{k_{l}+1}{2} \Bigg[\mu_{li}^{\{\tau\}} + \left\{ \frac{\sigma_l^2}{\sigma^2_0} + \frac{B_l^2}{\sigma^2_0(k_{l}+1)}\right\} \exp\left( -\mu_{li}^{\{\tau\}} + \frac{\sigma^{\{\tau\}2}_{li}}{2}\right) -1  + \log \frac{\sigma^2_0}{\sigma_l^2} \Bigg]\\
& +\frac{k_{L}+1}{2}\Bigg[\mu_{L}^{\{\tau\}} + \left\{\frac{\sigma_L^2}{\sigma^2_0} + \frac{B_L^2}{\sigma^2_0(k_{l}+1)}\right\} \exp\left( -\mu_{L}^{\{\tau\}} + \frac{\sigma^{\{\tau\}2}_{L}}{2} \right) -1  + \log \frac{\sigma^2_0}{\sigma_L^2} \Bigg] \\
& + \sum_{l=0}^{L} \sum_{i=1}^{k_{l+1}} \left[\frac{k_l+2}{2}(\log2-\mu_{li}^{\{\tau\}}-\mu^{\{\varsigma\}}) + \log \Gamma\left(\frac{k_l+2}{2}\right)   + \frac{1}{2} \exp\left(\mu_{li}^{\{\tau\}} + \frac{\sigma^{\{\tau\}2}_{li}}{2}+\mu^{\{\varsigma\}} + \frac{\sigma^{\{\varsigma\}2}}{2}\right) \right. \\
& \hspace{6cm} \left. - \log \sigma^{\{\tau\}}_{li} - \log \sqrt{2\pi} - \frac{1}{2} \right]\\
& + \log \Gamma(a_0)-a_0(b_0+\mu^{\{\varsigma\}})+b_0 \exp\left(\mu^{\{\varsigma\}}+
\frac{\sigma^{\{\varsigma\}}}{2}\right)-\log \sigma^{\{\varsigma\}} -\log \sqrt{2\pi}-\frac{1}{2}
\end{align*}
First line follows from Lemma A.4 in \cite{Jantre-et-al-2023}. The last inequality follows from $\sum_{j=1}^{k_l+1}{\overline{w}_{lij}^*}^2\leq B_l^2$ and \eqref{e:ass-lem-kl-ss-gl}. Note, $\sigma^2_0 = 1$, $\sigma_l^2 \leq 1$.  Let $\exp(\mu_{li}^{\{\tau\}}+{\sigma_{li}^{\{\tau\}}}^2/2)=(k_l+2)t_0''$ ,   $\exp(\sigma^2)=1+2/(k_l+2)$,   $\exp(\mu^{\{\varsigma\}}+{\sigma^{\{\varsigma\}}}^2/2)=1/t_0''$ and $\exp({\sigma^{\{\varsigma\}}}^2)=\exp(1/\log n)$. 

Let $b_l^*= (k_{l}+1)\log( k_{l+1} 2^{k_{l}+1})$,  $d_l^*=\prod_{m=0,m\neq L}^{L} B_m^2$, then $d_{\rm KL}(q^*,\pi) $ is bounded above by
\begin{align*}
 &\sum_{l=0}^{L-1} C n r_l + \sum_{l=0}^{L-1} \sum_{i=1}^{s^*_{l}} \frac{k_{l}+1}{2} \Bigg[\mu_{li}^{\{\tau\}} + \left\{ 1 + \frac{B_l^2}{(k_{l}+1)}\right\} \exp\left( -\mu_{li}^{\{\tau\}} + \frac{\sigma^{\{\tau\}2}_{li}}{2}\right) -1 - \log \sigma_l^2 \Bigg]\\
 &+\frac{k_{L}+1}{2}\Bigg[\mu_{L}^{\{\tau\}} + \left\{1 + \frac{B_L^2}{(k_{L}+1)} \right\} \exp\left(-\mu_{L}^{\{\tau\}} + \frac{\sigma^{\{\tau\}2}_{L}}{2} \right) - 1 - \log \sigma_L^2 \Bigg] \\
& + \sum_{l=0}^{L} \sum_{i=1}^{k_{l+1}} \left[\frac{k_l+2}{2}(\log2-\mu_{li}^{\{\tau\}}-\mu^{\{\varsigma\}}) + \log \Gamma\left(\frac{k_l+2}{2}\right)   + \frac{1}{2} \exp\left(\mu_{li}^{\{\tau\}} + \frac{\sigma^{\{\tau\}2}_{li}}{2}+\mu^{\{\varsigma\}} + \frac{\sigma^{\{\varsigma\}2}}{2}\right) \right. \\
& \hspace{6cm} \left. - \log \sigma^{\{\tau\}}_{li} - \log \sqrt{2\pi} - \frac{1}{2} \right]\\
& + \log \Gamma(a_0)-a_0(b_0+\mu^{\{\varsigma\}})+b_0 \exp\left(\mu^{\{\varsigma\}}+
\frac{\sigma^{\{\varsigma\}}}{2}\right)-\log \sigma^{\{\varsigma\}} -\log \sqrt{2\pi}-\frac{1}{2}\\
&\lesssim \sum_{l=0}^{L-1} Cnr_l+\sum_{l=0}^{L-1}\frac{s_l^*(k_l+1)}{2}\Big(\log (t_0''(k_l+2))+\Big\{1+\frac{B_l^2}{(k_l+1)}\Big\}\frac{1}{(k_l+2)t_0''}-\log \sigma_l^2\Big)\\
&+\frac{k_L+1}{2}\Big(\log (t_0''(k_L+2))+\Big\{1+\frac{B_L^2}{(k_L+1)}\Big\}\frac{1}{(k_L+2)t_0''}-\log \sigma_L^2\Big)+C_1 L +C_2 \log (\log n)\\
& \leq \sum_{l=0}^{L-1} C n r_l + \sum_{l=0}^{L-1} \frac{s^*_{l}(k_{l}+1)}{2} \Bigg[\log(t_0''(k_l+2)) + \Big(1+\frac{B_l^2}{(k_{l}+1)}\Big)\frac{1}{t_0''(k_l+2)} - \log \Bigg(\frac{s^*_{l}}{8n(L+1)4^{L-l}  b_l^*d_l^*} \Bigg) \Bigg] \\
& + \frac{k_{L}+1}{2}\Bigg[\log(t_0''(k_L+2)) +\Big(1+ \frac{B_L^2}{(k_{L}+1)}\Big)\frac{1}{t_0''(k_L+2)}  -\log \Bigg(\frac{1}{8n(L+1)b_L^*d_L^*}\Bigg)  \Bigg]  + C_1 L +C_2 \log (\log n)\\
& \lesssim \sum_{l=0}^L Cnr_l+ \sum_{l=0}^{L} \frac{s^*_{l}(k_{l}+1)}{2} \Bigg[-  \log\Big(\Big( \frac{B_l^2}{k_l+1}\Big)\frac{1}{{t_0''(k_l+2)}}\Big) + \Big(\frac{B_l^2}{(k_{l}+1)}\Big)\frac{1}{t_0''(k_l+2)} \\
& \hspace{6cm}+2(L+\log n+\sum \log B_m) \Bigg] \leq \sum_{l=0}^L Cnr_l
\end{align*}
\hfill $\blacksquare$
%----------------------------------------------------------------------

\section{Spike-and-Slab Group Horseshoe Proofs} 
\label{AppendixD}

\subsection{ELBO derivation}
\label{appendix:elbo-ssghs}
\vspace{-0.2in}
\begin{align*}
& \mathcal{L}= -\E_{q(\btheta,\bupvarpi)}[\log L(\btheta)] \\
&+ \sum_{l,j} q(\boldz_{lj} = 1) \int d_{\rm KL}(N(\bmu_{lj},\text{diag}(\bsigma^2_{lj})),N(0,\sigma_0^2 \widetilde{\tau}_{lj}^2\zeta^2\boldI))q( \beta_{lj})q(\alpha_{lj})q(\zeta_b)q(\zeta_a)d\beta_{lj}d\alpha_{lj}d\zeta_bd\zeta_b\\
&+\sum_{l,i} d_{\rm KL}(\text{Ber}(\gamma_{lj}),\text{Ber}(\lambda_l))+ \sum_{l,j} \left[d_{\rm KL}( LN(\mu^{\{\beta\}}_{lj}, {\sigma^{\{\beta\}}_{lj}}^2), IG\left(1/2,1\right)) +  d_{\rm KL}( LN(\mu^{\{\alpha\}}_{lj}, {\sigma^{\{\alpha\}}_{lj}}^2),  G\left(1/2,1\right)) \right] \\
& + d_{\rm KL}(LN(\mu^{\{\zeta_b\}}, {\sigma^{\{\zeta_b\}}}^2), IG(1/2,1)) + d_{\rm KL}(LN(\mu^{\{\zeta_a\}}, {\sigma^{\{\zeta_a\}}}^2), G(1/2,d_0^2))
\end{align*}

%----------------------------------------------------------------------

%----------------------------------------------------------------------
\subsection{Proof of Lemma 11.}

The proof of part-1 is same as SS-GL. We prove part-2 for SS-GHS below.

\begin{align*}
& \widetilde{\Pi}(\mathcal{F}(L,\bm{k},\bolds^\circ,\boldB^\circ)^c)  \leq \widetilde{\Pi}\left(\bigcup_{l=0}^L \{ ||\widetilde{\boldw}_l||_0>s_l^\circ\} \right) + \widetilde{\Pi}\left(\bigcup_{l=0}^L \{ ||\widetilde{\boldw}_l||_\infty > B_l^\circ \} \right) 
\end{align*}
\begin{align*}
\widetilde{\Pi}(\mathcal{F}(L,\bm{k},\bolds^\circ,\boldB^\circ)^c) & \leq \sum_{l=0}^L \widetilde{\Pi}(||\widetilde{\boldw}_l||_0>s_l^\circ) + \sum_{l=0}^L \widetilde{\Pi}(||\widetilde{\boldw}_l||_\infty > B_l^\circ) \\
& = \sum_{l=0}^L \int \int \sum_{\boldz}  \Pi(||\widetilde{\boldw}_l||_0>s_l^\circ|\btau^2,\zeta^2,\boldz) \pi(\boldz) \pi(\btau^2) \pi(\zeta^2) d\btau^2 d\zeta^2 \\
& \quad + \sum_{l=0}^L \int \int \sum_{\boldz} \Pi(||\widetilde{\boldw}_l||_\infty > B_l^\circ|\btau^2,\zeta^2,\boldz) \pi(\boldz) \pi(\btau^2) \pi(\zeta^2) d\btau^2 d\zeta^2\\
& \leq  \sum_{l=0}^L \prob(\sum_{i=1}^{k_{l+1}} z_{li}>s_{l}^\circ) + \sum_{l=0}^L \int \int \prob( \sup_{i=1,\cdots,k_{l+1}} ||\overline{\boldw}_{li}||_2 > B_l^\circ\Big|\btau^2,\zeta^2)\pi(\btau^2) \pi(\zeta^2) d\btau^2 d\zeta^2
\end{align*}

\noindent \underline{\it Part 2.}
\begin{flalign*}
& \sum_{l=0}^L \int \prob\left( \sup_{i=1,\cdots,k_{l+1}} ||\boldw_{li}||_2 > B_l^\circ\Big|\btau^2, \zeta^2\right)\pi(\btau^2)d\btau^2 \pi(\zeta^2) d\zeta^2\\
& \leq \sum_{l=0}^L \sum_{i=1}^{k_{l+1}} \int \int \prob\left( ||\boldw_{li}||_2 > B_l^\circ | \tau_{li}^2, \zeta^2 \right) \pi(\tau_{li}^2) d\tau_{li}^2 \hspace{1mm} \pi(\zeta^2) d\zeta^2\\ 
& \leq \sum_{l=0}^L \sum_{i=1}^{k_{l+1}} \int \int \prob\left( ||\boldw_{li}||_\infty > \frac{B_l^\circ}{k_{l}+1}\Big|\tau_{li}^2, \zeta^2 \right) \pi(\tau_{li}^2) d\tau_{li}^2 \hspace{1mm} \pi(\zeta^2) d\zeta^2\\ 
&\leq \sum_{l=0}^L \sum_{i=1}^{k_{l+1}} \int \int \sum_{j=1}^{k_l+1}\prob\left(|w_{lij}|>\frac{B_l^\circ}{k_{l}+1}\Big|\tau_{li}^2, \zeta^2 \right) \pi(\tau_{li}^2) d\tau_{li}^2 \hspace{1mm} \pi(\zeta^2) d\zeta^2 \\ 
&\leq 2\sum_{l=0}^L \sum_{i=1}^{k_{l+1}} \int \int \sum_{j=1}^{k_l+1} \exp\left(-\frac{{B_l^\circ}^2 (c^2_{\rm reg}+ \tau_{li}^2 \zeta^2)}{2(k_l+1)^2 c^2_{\rm reg}\tau_{li}^2 \zeta^2} \right) \pi(\tau_{li}^2) d\tau_{li}^2 \hspace{1mm} \pi(\zeta^2) d\zeta^2 \\ &\leq 2 \sum_{l=0}^L \sum_{i=1}^{k_{l+1}} \sum_{j=1}^{k_l+1}  \exp\left(-\frac{{ B_l^\circ}^2}{2c^2_{\rm reg}(k_l+1)^2}\right)
\leq \sum_{l=0}^L \sum_{i=1}^{k_{l+1}} \sum_{j=1}^{k_l+1} \frac{1}{(L+1) k_{l+1}(k_l+1)}\exp(-n\epsilon_n^2) 
\end{flalign*}
where the fourth inequality holds because $|w_{lij}|$ given $\tau_{li}^2$ and $\zeta^2$ is bound above by a $\Big|N(0,\frac{c^2_{\rm reg} \tau_{li}^2 \zeta^2 \sigma_0^2}{c^2_{\rm reg} + \tau_{li}^2 \zeta^2})\Big|$ random variable and we take $\sigma_0^2=1$ w.l.o.g. The above proof holds since $(B_l^\circ)^2=2(n\epsilon_n^2+\log (k_l+1)+\log k_{l+1}+\log(L+1)+\log 2)(k_l+1)^2c^2_{\rm reg}.$
\hfill $\blacksquare$

\subsection{Proof of Lemma 12 Condition 1.}
\begin{align}
\label{e:ass-lem-kl-ss-ghs}
{\it Assumption:}-\log \lambda_l =O\{(k_l+1) \vartheta_l\},\: -\log(1-\lambda_l) = O\{ (s_l/k_{l+1})(k_l+1) \vartheta_l\}
\end{align}
The initial parts of the proof are the same as the SS-GL and below we detail the remaining parts of the proof which differ from SS-GL in the SS-GHS case.
\vspace{2mm}

\noindent Let $\delta_n = (\sqrt{\sum r_l}B_l)/((L+1)(\prod_{j=0}^L B_j))$ and  $A = \{\overline{\boldw}_{li} :\enskip  ||\overline{\boldw}_{li} - \overline{\boldw}_{li}^*||_2 \leq \delta_n \}$
\begin{align*}
\widetilde{\Pi}\left(\calM_{\sqrt{\sum r_l}}\right)  &  =  \int \int \sum_{\boldz}\Pi\left(\calM_{\sqrt{\sum r_l}}\Big|\btau^2, \zeta^2, \boldz \right) \pi(\boldz) \pi(\btau^2) \pi(\zeta^2) d\btau^2 d\zeta^2\\
&\geq \int \int  \sum_{\{\boldz: z_{li}=1, i \in \mathcal{S}_l, z_{li}=0, i \in \mathcal{S}_l^c, l=0,\cdots, L\}} \Pi\left(\calM_{\sqrt{\sum r_l}}\Big|\btau^2, \zeta^2, \boldz \right)\pi(\boldz) \pi(\btau^2) \pi(\zeta^2) d\btau^2 d\zeta^2\\
&= \prod_{l=0}^L  (1-\lambda_l)^{k_{l+1}-s_l} \lambda_l^{s_l} \int \prod_{i=1}^{k_{l+1}} \int \E (\indicator_{\{\overline{\boldw}_{li} \in A \}}|\tau^2_{li},\zeta^2,z_{li}=1) \pi(\tau^2_{li}) \pi(\zeta^2) d\tau^2_{li} d \zeta^2 \\
&= \prod_{l=0}^L (1-\lambda_l)^{k_{l+1}-s_l} \lambda_l^{s_l} \int_0^\infty \prod_{i \in \mathcal{S}_l} \int_0^\infty \int_{\overline{\boldw}_{li} \in A} \Bigg[ \left(\frac{c^2_{\rm reg}+\tau^2_{li} \zeta^2}{2\pi c^2_{\rm reg} \tau^2_{li} \zeta^2}\right)^{\frac{k_l+1}{2}} \\
& \quad \times \prod_{j=1}^{k_l+1} \exp \left(- \frac{\overline{w}_{lij}^2(c^2_{\rm reg}+\tau^2_{li} \zeta^2)}{2 c^2_{\rm reg} \tau^2_{li} \zeta^2} \right) d \overline{w}_{lij} \pi(\tau_{li}^2) \pi(\zeta^2) d\tau^2_{li} d \zeta^2 \\
& \geq \prod_{l=0}^L \Bigg\{ (1-\lambda_l)^{k_{l+1}-s_l} \lambda_l^{s_l} \int_0^\infty \Bigg[ \prod_{i \in \mathcal{S}_l} \left(\frac{1+\tau_{li}^2\zeta^2/c^2_{\rm reg}}{2\pi}\right)^{\frac{k_l+1}{2}} \int_0^\infty  \Bigg[ \int_{\overline{w}^*_{lij}-\frac{\delta_n}{\sqrt{k_l+1}}}^{\overline{w}^*_{lij}+\frac{\delta_n}{\sqrt{k_l+1}}} \prod_{j=1}^{k_l+1}\\
&\exp \left(- \frac{\overline{w}_{lij}^2 (c^2_{\rm reg} + \tau^2_{li} \zeta^2)}{2 c^2_{\rm reg} \tau^2_{li} \zeta^2} \right) d \overline{w}_{lij}\Bigg]K (\tau^2_{li})^{-\frac{k_l+1}{2}-a-1} L(\tau^2_{li}) d\tau^2_{li} \Bigg] K' (\zeta^2)^{-\frac{s_l(k_l+1)}{2}-a'-1} L(\zeta^2) d \zeta^2 \Bigg\} \\
& \geq \prod_{l=0}^L \Bigg\{ (1-\lambda_l)^{k_{l+1}-s_l} \lambda_l^{s_l} \Bigg[ \prod_{i \in \mathcal{S}_l} \left(\frac{1+t_0t_0'/c^2_{\rm reg}}{2\pi}\right)^{\frac{k_l+1}{2}} c_0 K \left(\frac{2\delta_n}{\sqrt{k_l+1}}\right)^{k_l+1} \\
&  \exp \left(- \sum_{j=1}^{k_l+1} \widehat{w}_{lij}^2\left(\frac{1}{c^2_{\rm reg}}+\frac{1}{t_0t_0'}\right) \right) \int_{t_0}^\infty (\tau^2_{li})^{-\frac{k_l+1}{2}-a-1} d\tau^2_{li} \Bigg] \hspace{3mm} K' c_0' \int_{t_0'}^\infty (\zeta^2)^{-\frac{s_l(k_l+1)}{2}-a'-1} d \zeta^2 \Bigg\} \\
&= \prod_{l=0}^L \Bigg\{ (1-\lambda_l)^{k_{l+1}-s_l} \lambda_l^{s_l} \int_0^\infty \Bigg[ \prod_{i \in \mathcal{S}_l} \left(\frac{1+t_0t_0'/c^2_{\rm reg}}{2\pi}\right)^{\frac{k_l+1}{2}}\left(\frac{2\delta_n} {\sqrt{k_l+1}}\right)^{k_l+1}Kc_0 \\ 
& \quad \exp \left(- \sum_{j=1}^{k_l+1} \widehat{w}_{lij}^2\left(\frac{1}{c^2_{\rm reg}} +\frac{1}{t_0t_0'}\right)\right) \frac{t_0^{-\frac{k_l+1}{2}-a} }{\frac{k_l+1}{2}+a}  \Bigg] K'c_0' \frac{ {t_0'}^{-\frac{s_l(k_l+1)}{2}-a'} }{\frac{s_l(k_l+1)}{2}+a'} \Bigg\}\\
& \nonumber = \exp\Bigg[ -\sum_{l=0}^L\Bigg\{ s_l \log \left( \frac{1}{\lambda_l}\right)+(k_{l+1}-s_l)\log\left(\frac{1}{1-\lambda_l}\right) - C'-\frac{s_l(k_l+1)}{2} \log\left(1+\frac{t_0t_0'}{c^2_{\rm reg}}\right)\\
&\nonumber  +\left(\frac{s_l(k_l+1)}{2}+a'\right) \log t_0' + \log \left(\frac{s_l(k_l+1)}{2}+a'\right) + \frac{s_l(k_l+1)}{2} \log 2\pi +\sum_{i\in\mathcal{S}_l} \Bigg(-C\\
&\nonumber  - (k_l+1) \log \frac{2\delta_n}{\sqrt{k_l+1}}+  \log t_0^{\frac{(k_l+1)}{2}+a} + \log \left(\frac{(k_l+1)}{2}+a\right)  + \sum_{j=1}^{k_l+1} \widehat{w}_{lij}^2\left(\frac{1}{t_0t_0'}+\frac{1}{c^2_{\rm reg}}\right) \Bigg) \Bigg\}  \Bigg] 
\end{align*}
where the third equality follows $\E (\indicator_{\{\overline{\boldw}_{li} \in A \}}|\tau_{li}^2,\zeta^2,z_{li}=0)=1 $ since $||\overline{\boldw}_{li}^*||_2=0$, for $i \in \mathcal{S}_l^c$. The last equality is by mean value theorem and $\widehat{w}_{lij} \in [\overline{w}^*_{lij}-\delta_n/\sqrt{k_l+1},\overline{w}^*_{lij}+\delta_n/\sqrt{k_l+1}]$.

\begin{align}
& \nonumber \gtrsim  \exp\Bigg[ -\sum_{l=0}^L\Bigg\{ s_l \log \left( \frac{1}{\lambda_l}\right)+(k_{l+1}-s_l)\log\left(\frac{1}{1-\lambda_l}\right)  -\frac{s_l(k_l+1)}{2} \log\left(1+\frac{t_0t_0'}{c^2_{\rm reg}}\right)\\ 
& \nonumber  +\left(\frac{s_l(k_l+1)}{2}+a' \right) \log t_0'+ \log \left(\frac{s_l(k_l+1)}{2}+a' \right) + \frac{s_l(k_l+1)}{2}\left(\log 2\pi - 2 \log \frac{2\delta_n}{\sqrt{k_l+1}}\right)\\
& \nonumber  + s_l\left(\frac{k_l+1}{2}+a\right) \log t_0 + s_l\log \left(\frac{k_l+1}{2}+a\right)+ \sum_{i \in \mathcal{S}_l}\sum_{j=1}^{k_l+1} \widehat{w}_{lij}^2\left(\frac{1}{t_0t_0'}+\frac{1}{c^2_{\rm reg}}\right) \Bigg\} \Bigg] \\
& \nonumber \gtrsim \exp\Bigg[ -\sum_{l=0}^L\Bigg\{ s_l \log \left( \frac{1}{\lambda_l}\right)+(k_{l+1}-s_l)\log\left(\frac{1}{1-\lambda_l}\right) -  \frac{s_l(k_l+1)}{2} \log \left(\frac{1}{t_0t_0'}+\frac{1}{c^2_{\rm reg}}\right)\\
& \qquad + \frac{s_l(k_l+1)}{2} \Big( \log (k_l+1) - \log (\delta_n^2) \Big) + \sum_{i \in \mathcal{S}_l} \sum_{j=1}^{k_l+1} \widehat{w}_{lij}^2\left(\frac{1}{t_0t_0'}+\frac{1}{c^2_{\rm reg}}\right) \Bigg\} \Bigg]
\label{ss_ghs_KL_lemma_main_expression}
\end{align}
Here, $\sum_{l=0}^L  \sum_{i\in \mathcal{S}_l} \sum_{j=1}^{k_l+1} \widehat{w}_{lij}^2 (1/{t_0t_0'}+1/{c^2_{\rm reg}})$  is bounded above by
\begin{flalign}
\label{ss_ghs_KL_lemma_weight_bound}
\nonumber &  \left(\frac{1}{t_0t_0'}+\frac{1}{c^2_{\rm reg}}\right)\sum_{l=0}^L \sum_{i\in \mathcal{S}_l} \sum_{j=1}^{k_l+1} \max ((\overline{w}^*_{lij}-\frac{\delta_n}{\sqrt{k_l+1}})^2,(\overline{w}^*_{lij}+\frac{\delta_n}{\sqrt{k_l+1}})^2) \\
\nonumber & \leq 2\left(\frac{1}{t_0t_0'}+\frac{1}{c^2_{\rm reg}}\right) \sum_{l=0}^L \sum_{i\in \mathcal{S}_l} \sum_{j=1}^{k_l+1}(\overline{w}_{lij}^{*2} + \delta_n^2/(k_{l}+1))  \\
\nonumber & \leq \sum_{l=0}^L \sum_{i\in \mathcal{S}_l} 2||\overline{\boldw}_{li}^*||_2^2\left(\frac{1}{t_0t_0'}+\frac{1}{c^2_{\rm reg}}\right) +\sum_{l=0}^L \sum_{i\in \mathcal{S}_l} \frac{2\delta_n^2}{(k_l+1)}\left(\frac{1}{t_0t_0'}+\frac{1}{c^2_{\rm reg}}\right) \\
& \leq \sum_{l=0}^L 2s_l(k_l+1)\left(\frac{B_l^2}{k_l+1} +\frac{1}{(k_l+1)}\right)\left(\frac{1}{t_0t_0'}+\frac{1}{c^2_{\rm reg}}\right), \text{since $\delta_n \to 0$.}
\end{flalign}
\begin{flalign}
\nonumber & \sum_{l=0}^L \Bigg( s_l \log \left( \frac{1}{\lambda_l}\right)+(k_{l+1}-s_l)\log\left(\frac{1}{1-\lambda_l}\right)  -\frac{s_l(k_l+1)}{2}\Bigg(\log \left(\frac{1}{t_0t_0'}+\frac{1}{c^2_{\rm reg}}\right)-\log \frac{k_l+1}{ \delta_n^2} \Bigg)\\
 & \leq \sum_{l=0}^L Cnr_l+\sum_{l=0}^L \frac{s_l(k_l+1)}{2}\Bigg\{-\log\left( \frac{B_l^2}{k_l+1}\left(\frac{1}{t_0t_0'}+\frac{1}{c^2_{\rm reg}}\right)\right) +\log ((L+1)n)+ \sum \log B_m^2  \Bigg\}\nonumber\\
& \leq \sum_{l=0}^L Cnr_l
 \label{ss_ghs_KL_lemma_rest_bound}
\end{flalign}
where the first inequality follows from \eqref{e:ass-lem-kl-ss-ghs} and expanding $\delta_n$. The last inequality follows since $n\sum r_l \to \infty$. Combining \eqref{ss_ghs_KL_lemma_weight_bound} and \eqref{ss_ghs_KL_lemma_rest_bound} and replacing \eqref{ss_ghs_KL_lemma_main_expression}, the proof follows.
\hfill $\blacksquare$

\subsection{Proof of Lemma 12 Condition 2.}
\begin{align*}
{\it Assumption:}-\log \lambda_l =O\{(k_l+1) \vartheta_l\},\: -\log(1-\lambda_l) = O\{ (s_l/k_{l+1})(k_l+1) \vartheta_l\}
\end{align*}
Initial parts of proof are same as SS-GL  and below is the remaining  proof which differs in SS-GHS. Let $\tau_{li}^2=\beta_{li}\alpha_{li}$, $\zeta^2=\zeta_a\zeta_b$, $\tau_L^2=\beta_L \alpha_L$. 

We can therefore upperbound $d_{\rm KL}(q^*,\pi)$ as follows
\begin{align*}
d_{\rm KL}(q^*,\pi) & \leq \log\frac{1}{\pi(\boldz^*)} + \bigintssss \hspace{-1mm} \bigintssss \Bigg[ \indicator (\boldz=\boldz^*) \sum_{l,i} \bigintssss \hspace{-1mm} \bigintssss d_{\rm KL}\Big\{\prod_{l=0}^{L-1} \prod_{i=1}^{k_{l+1}} \prod_{j=1}^{k_{l}+1} \Big\{ z_{li}\calN(\overline{w}_{lij}^*,\sigma_l^2) +(1-z_{li})\delta_0 \Big\} \\
&   \prod_{j=1}^{k_{L}+1} \calN (\overline{w}_{Lj}^*,\sigma_L^2) \Big\} ,\Big\{ \prod_{l=0}^{L-1} \prod_{i=1}^{k_{l+1}} \prod_{j=1}^{k_{l}+1} \Big\{ z_{li}\calN (0, \frac{ \sigma^2_0c^2_{\rm reg} \tau_{li}^2 \zeta^2}{c^2_{\rm reg} + \tau_{li}^2 \zeta^2} ) +(1-z_{li})\delta_0 \Big\} \prod_{j=1}^{k_{L}+1} \calN (0, \frac{\sigma^2_0 c^2_{\rm reg}\tau_L^2\zeta^2 }{c^2_{\rm reg}+\tau_L^2\zeta^2 }) \Big\}\\
& \phantom{\bigintss} \times LN(\mu^{\{\beta\}}_{li}, {\sigma^{\{\beta\}}_{li}}^2) LN(\mu^{\{\alpha\}}_{li}, {\sigma^{\{\alpha\}}_{li}}^2) d \beta_{li} d \alpha_{li} \Bigg] LN(\mu^{\{\zeta_b\}}, {\sigma^{\{\zeta_b\}}}^2) LN(\mu^{\{\zeta_a\}}, {\sigma^{\{\zeta_a\}}}^2) d \zeta_b d \zeta_a\\
&+ \sum_{l,i} \Bigg[ d_{\rm KL} \left( LN\left(\mu_{li}^{\{\beta\}}, \sigma^{\{\beta\}2}_{li}\right),  IG \left(\frac{1}{2},1\right)\right) + d_{\rm KL}\left( LN\left(\mu_{li}^{\{\alpha\}},\sigma^{\{\alpha\}2}_{li}\right),  \Gamma \left(\frac{1}{2},1\right)\right) \Bigg] \\
& + d_{\rm KL} \left( LN\left(\mu^{\{\zeta_b\}}, \sigma^{\{\zeta_b\}2}\right),  IG \left(\frac{1}{2},1\right)\right) + d_{\rm KL}\left( LN\left(\mu^{\{\zeta_a\}},\sigma^{\{\zeta_a\}2}\right),  \Gamma \left(\frac{1}{2},d_0^2\right)\right) \\
& = \sum_{l=0}^{L-1} \Bigg(s^*_{l} \log \frac{1}{\lambda_l} + (k_{l+1}-s^*_{l}) \log \frac{1}{1-\lambda_l} \Bigg) \\
& + \sum_{l=0}^{L-1} \sum_{i=1}^{k_{l+1}} \sum_{j=1}^{k_{l}+1} z_{li}^* \Bigg\{\frac{1}{2}\log \frac{c^2_{\rm reg}\sigma^2_0}{\sigma_l^2} + \frac{(\mu_{li}^{\{\beta\}} + \mu_{li}^{\{\alpha\}} + \mu^{\{\zeta_b\}} + \mu^{\{\zeta_a\}})}{2} - \frac{\log (c^2_{\rm reg}+\widehat{\beta}_{li}\widehat{\alpha}_{li}\widehat{\zeta}_b \widehat{\zeta}_a)}{2} - \frac{1}{2} \\ 
&+ \frac{\sigma_l^2+{\overline{w}_{lij}^{*2}}}{2\sigma^2_0}[\exp( -\mu_{li}^{\{\beta\}} -\mu_{li}^{\{\alpha\}} - \mu^{\{\zeta_b\}} - \mu^{\{\zeta_a\}}  + \frac{\sigma^{\{\beta\}2}_{li} + \sigma^{\{\alpha\}}_{li} + \sigma^{\{\zeta_b\}2} + \sigma^{\{\zeta_a\}2}}{2}) + \frac{1}{c^2_{\rm reg}} ] \Bigg\} \\
&+ \sum_{j=1}^{k_{L}+1} \Bigg\{\frac{1}{2}\log \frac{c^2_{\rm reg}\sigma^2_0}{\sigma_L^2} + \frac{(\mu_{L}^{\{\beta\}} + \mu_{L}^{\{\alpha\}} + \mu^{\{\zeta_b\}} + \mu^{\{\zeta_a\}})}{2} - \frac{\log (c^2_{\rm reg}+\widehat{\beta}_{L}\widehat{\alpha}_{L}\widehat{\zeta}_b \widehat{\zeta}_a)}{2} - \frac{1}{2} \\
&  + \frac{\sigma_L^2+{\overline{w}_{Lj}^{*2}}}{2\sigma^2_0} [ \exp ( -\mu_{L}^{\{\beta\}} - \mu_{L}^{\{\alpha\}} -\mu^{\{\zeta_b\}} - \mu^{\{\zeta_a\}}  + \frac{\sigma^{\{\beta\}2}_{L} + \sigma^{\{\alpha\}2}_{L} + \sigma^{\{\zeta_b\}2} + \sigma^{\{\zeta_a\}2}}{2} ) + \frac{1}{c^2_{\rm reg}} ]  \Bigg\} \\
& + \sum_{l=0}^{L} \sum_{i=1}^{k_{l+1}} \Bigg[ \frac{1}{2} \mu_{li}^{\{\beta\}} + \exp\left( - \mu_{li}^{\{\beta\}} + \frac{\sigma^{\{\beta\}2}_{li}}{2}\right) - \log \sigma^{\{\beta\}}_{li} - \log \sqrt{2} - \frac{1}{2} \Bigg] \\
&  + \sum_{l=0}^{L} \sum_{i=1}^{k_{l+1}} \Bigg[ - \frac{1}{2} \mu_{li}^{\{\alpha\}} + \exp\left(\mu_{li}^{\{\alpha\}} + \frac{\sigma^{\{\alpha\}2}_{li}}{2}\right) - \log \sigma^{\{\alpha\}}_{li} - \log \sqrt{2} - \frac{1}{2} \Bigg] \\
& + \frac{1}{2} \mu^{\{\zeta_b\}} + \exp\left( - \mu^{\{\zeta_b\}} + \frac{\sigma^{\{\zeta_b\}2}}{2}\right) - \log \sigma^{\{\zeta_b\}} - \log \sqrt{2} - \frac{1}{2} \\
& + \log d_0 - \frac{1}{2} \mu^{\{\zeta_a\}} + \frac{1}{d_0^2}\exp\left(\mu^{\{\zeta_a\}} + \frac{\sigma^{\{\zeta_a\}2}}{2}\right) - \log \sigma^{\{\zeta_a\}} - \log \sqrt{2} - \frac{1}{2} 
\end{align*}
Simplifying further, we get
\begin{align*}
d_{\rm KL}(q^*,\pi) & \leq \sum_{l=0}^{L-1} C n r_l +
\sum_{l=0}^{L} \frac{s_l^*(k_{l}+1)}{2} \Bigg\{ (\mu_{li}^{\{\beta\}} + \mu_{li}^{\{\alpha\}} + \mu^{\{\zeta_b\}} + \mu^{\{\zeta_a\}}) + \log \frac{\sigma^2_0}{\sigma_l^2}  + \Bigg\{ \frac{\sigma_l^2}{\sigma^2_0} + \frac{B_l^2}{\sigma^2_0(k_{l}+1)} \Bigg\} \\
& \times \Bigg[ \exp \Bigg( -(\mu_{li}^{\{\beta\}} + \mu_{li}^{\{\alpha\}} + \mu^{\{\zeta_b\}} + \mu^{\{\zeta_a\}}) + \frac{\sigma^{\{\beta\}2}_{li} + \sigma^{\{\alpha\}2}_{li} + \sigma^{\{\zeta_b\}2} + \sigma^{\{\zeta_a\}2}}{2}\Bigg)  + \frac{1}{c^2_{\rm reg}} \Bigg] \Bigg\}\\
&-\sum_{l=0}^{L-1} \frac{s_l^*(k_l+1)}{2}\log \left(1+\frac{\widehat{\beta}_{li}\widehat{\alpha}_{li}\widehat{\zeta}_b \widehat{\zeta}_a}{c^2_{\rm reg}}\right)-\frac{k_L+1}{2}\log \left(1+\frac{\widehat{\beta}_{L}\widehat{\alpha}_{L}\widehat{\zeta}_b \widehat{\zeta}_a}{c^2_{\rm reg}}\right)\\
&  + \sum_{l=0}^{L} \sum_{i=1}^{k_{l+1}} \Bigg[ \frac{1}{2} \mu_{li}^{\{\beta\}} + \exp\left( - \mu_{li}^{\{\beta\}} + \frac{\sigma^{\{\beta\}2}_{li}}{2}\right) - \log \sigma^{\{\beta\}}_{li} - \log \sqrt{2} - \frac{1}{2} \Bigg] \\
&  + \sum_{l=0}^{L} \sum_{i=1}^{k_{l+1}} \Bigg[ - \frac{1}{2} \mu_{li}^{\{\alpha\}} + \exp\left(\mu_{li}^{\{\alpha\}} + \frac{\sigma^{\{\alpha\}2}_{li}}{2}\right) - \log \sigma^{\{\alpha\}}_{li} - \log \sqrt{2} - \frac{1}{2} \Bigg] \\
& \quad + \frac{1}{2} \mu^{\{\zeta_b\}} + \exp\left( - \mu^{\{\zeta_b\}} + \frac{\sigma^{\{\zeta_b\}2}}{2}\right) - \log \sigma^{\{\zeta_b\}} - \log \sqrt{2} - \frac{1}{2} \\
& \quad - \frac{1}{2} \mu^{\{\zeta_a\}} + \exp\left(\mu^{\{\zeta_a\}} + \frac{\sigma^{\{\zeta_a\}2}}{2}\right) - \log \sigma^{\{\zeta_a\}} - \log \sqrt{2} - \frac{1}{2}\\
&\leq \sum_{l=0}^{L-1} C n r_l + 
\sum_{l=0}^{L}\frac{s_l^*(k_{l}+1)}{2} \Bigg\{(\mu_{li}^{\{\beta\}} + \mu_{li}^{\{\alpha\}} + \mu^{\{\zeta_b\}} + \mu^{\{\zeta_a\}})  + \Bigg\{ \sigma_l^2 + \frac{B_l^2}{(k_{l}+1)} \Bigg\} \\
&  \times \Bigg[ \exp ( -\mu_{li}^{\{\beta\}} - \mu_{li}^{\{\alpha\}} - \mu^{\{\zeta_b\}} - \mu^{\{\zeta_a\}} + \frac{\sigma^{\{\beta\}2}_{li} + \sigma^{\{\alpha\}2}_{li} + \sigma^{\{\zeta_b\}2} + \sigma^{\{\zeta_a\}2}}{2}) + \frac{1}{c^2_{\rm reg}} \Bigg]  - \log \sigma_l^2\Bigg\}\\
&-\sum_{l=0}^{L-1} \frac{s_l^*(k_l+1)}{2}\log \left(1+\frac{\widehat{\beta}_{li}\widehat{\alpha}_{li}\widehat{\zeta}_b \widehat{\zeta}_a}{c^2_{\rm reg}}\right)-\frac{k_L+1}{2}\log \left(1+\frac{\widehat{\beta}_{L}\widehat{\alpha}_{L}\widehat{\zeta}_b \widehat{\zeta}_a}{c^2_{\rm reg}}\right)\\
&  + \sum_{l=0}^{L} \sum_{i=1}^{k_{l+1}} \Bigg[ \frac{1}{2} \mu_{li}^{\{\beta\}} + \exp\left( - \mu_{li}^{\{\beta\}} + \frac{\sigma^{\{\beta\}2}_{li}}{2}\right) - \log \sigma^{\{\beta\}}_{li} - \log \sqrt{2} - \frac{1}{2} \Bigg] \\
& + \sum_{l=0}^{L} \sum_{i=1}^{k_{l+1}} \Bigg[ - \frac{1}{2} \mu_{li}^{\{\alpha\}} + \exp\left(\mu_{li}^{\{\alpha\}} + \frac{\sigma^{\{\alpha\}2}_{li}}{2}\right) - \log \sigma^{\{\alpha\}}_{li} - \log \sqrt{2} - \frac{1}{2} \Bigg] \\
&  + \frac{1}{2} \mu^{\{\zeta_b\}} + \exp\left( - \mu^{\{\zeta_b\}} + \frac{\sigma^{\{\zeta_b\}2}}{2}\right) - \log \sigma^{\{\zeta_b\}} - \log \sqrt{2} - \frac{1}{2} \\
&- \frac{1}{2} \mu^{\{\zeta_a\}} + \exp\left(\mu^{\{\zeta_a\}} + \frac{\sigma^{\{\zeta_a\}2}}{2}\right) - \log \sigma^{\{\zeta_a\}} - \log \sqrt{2} - \frac{1}{2} 
\end{align*}
The first line follows by Lemma 
 A.4 in \cite{Jantre-et-al-2023}. We also use $\sum_{j=1}^{k_l+1}{\overline{w}_{lij}^*}^2\leq B_l^2$, \eqref{e:ass-lem-kl-ss-ghs}, $\sigma^2_0 = 1$, $\sigma_l^2 \leq 1$. Further, we let $\exp( - \mu_{li}^{\{\beta\}}+\sigma^{\{\beta\}2}_{li}/2)  =1/t_0^2$, $\sigma^{\{\beta\}2}_{li}=1/\log n$, $\exp( - \mu_{li}^{\{\alpha\}}+\sigma^{\{\alpha\}2}_{li}/2) =t_0$, $\sigma^{\{\alpha\}2}_{li}=1/\log n$, $\exp( - \mu^{\{\zeta_b\}}+\sigma^{\{\zeta_b\}2}/2) =1/{t_0'}^2$, $\sigma^{\{\zeta_b\}2}=1/\log n$. Finally, we let $\exp( - \mu^{\{\zeta_a\}}+\sigma^{\{\zeta_a\}2}/2)  =t_0'$, $\sigma^{\{\zeta_a\}2}=1/\log n$.
\begin{align*}
    &\lesssim \sum_{l=0}^{L-1}Cnr_l+\sum_{l=0}^L \frac{s_l^*(k_l+1)}{2}\Bigg\{-\log \Big(\frac{1}{t_0t_0'}+\frac{1}{c^2_{\rm reg}}\Big)-\log \sigma_l^2+\Big(1+\frac{B_l^2}{k_l+1}\Big)\Big(\frac{1}{t_0t_0'}+\frac{1}{c^2_{\rm reg}}\Big)\Bigg\}\\
&+\sum_{l=0}^L k_{l+1}\Bigg\{\log t_0+\frac{1}{4\log n}+\frac{1}{t_0^2}+\frac{1}{2}\log (\log n)-\log \sqrt{2}-\frac{1}{2}\Bigg\}\\
&+\sum_{l=0}^L k_{l+1}\Bigg\{\frac{1}{2}\log t_0-\frac{1}{4 \log n}+\frac{1}{t_0}+\frac{1}{2}\log (\log n)-\log \sqrt{2}-\frac{1}{2}\Bigg\}\\
&+\Bigg\{\log t_0'+\frac{1}{4\log n}+\frac{1}{t_0'^2}+\frac{1}{2}\log (\log n)-\log \sqrt{2}-\frac{1}{2}\Bigg\}\\
&+\Bigg\{\frac{1}{2}\log t_0'-\frac{1}{4 \log n }+\frac{1}{t_0'}+\frac{1}{2}\log(\log n)-\log \sqrt{2}-\frac{1}{2}\Bigg\}\\
&\lesssim \sum_{l=0}^{L-1} C n r_l + \frac{1}{2}\sum_{l=0}^{L} s_l^*(k_{l}+1) \Big( -\log \Big(\frac{B_l^2}{k_l+1}\Big(\frac{1}{t_0t_0'}+\frac{1}{c^2_{\rm reg}}\Big)\Big)+ \Big(\frac{B_l^2}{k_l+1}\Big)\Big(\frac{1}{t_0t_0'}+\frac{1}{c^2_{\rm reg}}\Big) \\
& \hspace{6cm} +2L+2\log n+2\sum \log B_m\Big)\leq \sum_{l=0}^L Cnr_l
\end{align*}
\hfill $\blacksquare$

\end{document}